%% file: main.tex
\documentclass[10pt]{article}
\usepackage[utf8]{inputenc}
\usepackage{float}
\usepackage{graphicx}
\usepackage{subfig}
\usepackage{makecell}
\usepackage{adjustbox}
\usepackage{amsfonts}
\usepackage{blkarray, bigstrut}
\usepackage{svg}
\usepackage{enumitem}
\usepackage{authblk}
\usepackage{amsmath}
\usepackage{rotating}
\DeclareMathOperator*{\argmax}{\arg,\max}
\usepackage{hyperref}
\newcolumntype{C}{>{\centering\arraybackslash}X}
\usepackage{tabularx,booktabs}
\title{Sensor Data for Human Activity Recognition: Feature Representation and Benchmarking}

\author[1]{Fl\'avia Alves}
\author[1]{Martin Gairing}

\author[2]{Frans A. Oliehoek}
\author[1]{\\Thanh-Toan Do}

\affil[1]{Department of Computer Science, University of Liverpool, United Kingdom; e-mail: F.Alves@liverpool.ac.uk, gairing@liverpool.ac.uk, Thanh-Toan.Do@liverpool.ac.uk}
\affil[2]{Department of Intelligent Systems, Delft University of Technology, Netherlands; e-mail: f.a.oliehoek@tudelft.nl}

\date{}

\begin{document}

\maketitle

\input{abstract}
\input{intro}
\input{relatedwork}
\input{preliminary}
\input{propose}
\input{experiment}
\input{conclusion}

\section*{Acknowledgements}
The support of NVIDIA Corporation is gratefully acknowledged with the donation of the Quadro P6000 GPU used for this research.

\bibliographystyle{abbrv}
\bibliography{bib}

\appendix
\input{appendix}

\end{document}

%% file: abstract.tex
\begin{abstract}
    
The field of Human Activity Recognition (HAR) focuses on obtaining and analysing data captured from monitoring devices (e.g. sensors). There is a wide range of applications within the field; for instance, assisted living, security surveillance, and intelligent transportation. In HAR, the development of Activity Recognition models is dependent upon the data captured by these devices and the methods used to analyse them, which directly affect performance metrics. In this work, we address the issue of accurately recognising human activities using different Machine Learning (ML) techniques. We propose a new feature representation based on consecutive occurring observations and compare it against previously used feature representations using a wide range of classification methods.
Experimental results demonstrate that techniques based on the proposed representation outperform the baselines and a better accuracy was achieved for both highly and less frequent actions. We also investigate how the addition of further features and their pre-processing techniques affect performance results leading to state-of-the-art accuracy on a Human Activity Recognition dataset.
    
\end{abstract}

%% file: intro.tex
\section{Introduction}

Over the past fifteen years, extensive research has been carried out in the field of Human Activity Recognition \cite{delahoz18}. This has been largely motivated by the technological advancement in monitoring devices within several research areas. One example where this applies is the improvement of services in elderly care. As discussed in \cite{kaye11}, any form of traditional methodology (e.g. in-person visits and telephone interviews) has its inherent limitations and a 24-hour continuous monitoring contributes towards mitigating the risks associated with them. Therefore, the potential that HAR has in order to detect physical and cognitive changes provides a great opportunity for the development of bespoke prevention plans.

HAR aims to infer the actions taken by an individual using monitoring sensors \cite{chen12}. A generic activity recognition model takes as input the data collected by the sensors and aims to accurately classify the activities of the individual.

We use the van Kasteren dataset \cite{kasteren11} which consists of binary sensor activity from three different houses (A, B and C) \cite{kasteren11}. The binary sensors capture human activity by indicating, for instance, if a door or a cupboard is open or closed, if the toilet is being flushed, or if a person is sitting on a couch, lying in bed or moving in a specific area. The dataset provides sensor readings in 60 second intervals.

In this paper, we present a thorough study of ML techniques including probabilistic (Na\"ive Bayes, Hidden Markov Model, Hidden Semi-Markov Model and Conditional Random Field) and neural network based (Recurrent Neural Network, Long Short-Term Memory Network, Gated Recurrent Unit, Multi-Layer Perceptron and a Long Short-Term Memory Network with a Conditional Random Field layer) models to the classification task. The main contributions are: (i) A new feature representation (observation-based) is proposed and compared against the state-of-the-art results for other feature representations. The proposed representation outperforms the others and, in general, is able to produce a better accuracy for both dominant and minor classes; (ii) We provide an extensive evaluation and analysis of the aforementioned classification models. Our analysis shows that the Conditional Random Field model performs best using an observation-based representation; (iii) Our best method produces state-of-the-art accuracy on the van Kasteren dataset.

%% file: relatedwork.tex
\subsection{Related Work}

A number of papers have proposed techniques for classifying the data in \cite{kasteren11} and evaluated them using two evaluation metrics: the overall accuracy and the mean per class accuracy\footnote{The accuracy calculates how often the predictions match the class labels and the mean per class accuracy calculates the average of the per-class accuracies.}.

Both generative (e.g. Na\"ive Bayes (NB)~\cite{kasteren11}, Hidden Markov Models (HMMs) \cite{kasteren08, kasteren11}) and discriminative (Support Vector Machines (SVMs)~\cite{arifoglu17}, Conditional Random Fields (CRFs)~\cite{kasteren08, kasteren11}) methods have been evaluated against this dataset. The state-of-the-art methods are Hidden Semi-Markov models (HSMMs) and CRFs~\cite{kasteren08, kasteren11} depending on which evaluation metric is being considered.

\begin{table}
    \caption{Accuracy and mean per class accuracy rates (\%) and their standard deviation for state-of-the-art methods obtained by reproducing results presented in \cite{kasteren11} and our best results for houses A, B and C}
    \begin{adjustbox}{width=\columnwidth,center}
    \small
    \label{sotares}
    \begin{tabularx}{\textwidth}{c*2{C}{C}}
        \toprule
        House & Model & Mean per class accuracy & Accuracy\\
        \midrule
        A & HSMM & 74.96 $\pm$ 12.1  & 91.81 $\pm$ 5.88\\
         & & (75.0 $\pm$ 12.1 \cite{kasteren11}) & (91.8 $\pm$ 5.9 \cite{kasteren11})\\ 
        A & CRF & 69.35 $\pm$ 12.07 & 96.93 $\pm$ 2.11\\
         & & (65.8 $\pm$ 14.0 \cite{kasteren11}) & (96.4 $\pm$ 2.4 \cite{kasteren11})\\
        A & This paper & \textbf{88.40} $\pm$ \textbf{12.43} & \textbf{98.95} $\pm$ \textbf{1.62}\\
        \midrule
        B & HSMM & 65.18 $\pm$ 13.41 & 82.27 $\pm$ 13.51\\
         & & (65.2 $\pm$ 13.4 \cite{kasteren11}) & (82.3 $\pm$ 13.5 \cite{kasteren11})\\
        B & CRF & 58.06 $\pm$ 7.01 & 94.99 $\pm$ 5.71\\
         & & (51.5 $\pm$ 8.5 \cite{kasteren11}) & (92.9 $\pm$ 6.2 \cite{kasteren11})\\
        B & This paper & \textbf{79.08} $\pm$ \textbf{22.35} & \textbf{96.07} $\pm$ \textbf{6.35}\\
        \midrule
        C & HSMM & 55.98 $\pm$ 15.4  & 84.48 $\pm$ 13.17\\
         & & (56.0 $\pm$ 15.4 \cite{kasteren11}) & (84.5 $\pm$ 13.2 \cite{kasteren11})\\
        C & CRF & 46.79 $\pm$ 15.63  & 90.69 $\pm$ 9.05\\
         & & (40.4 $\pm$ 16.0 \cite{kasteren11}) & (89.7 $\pm$ 8.4 \cite{kasteren11})\\
        C & This paper & \textbf{76.54} $\pm$ \textbf{18.99} & \textbf{94.10} $\pm$ \textbf{15.27}\\
        \bottomrule
    \end{tabularx}
    \end{adjustbox}
\end{table}

From the literature we are able to identify the state-of-the-art methods which provide the best accuracy and mean per class accuracy, in particular, CRFs and HSMMs, respectively. The previous best results for those metrics and their standard deviation and our improved results are summarised in Table~\ref{sotares}. Our results for HSMM and CRF differ from the ones that were published in \cite{kasteren11}, in particular, the values of the mean per class accuracy that we obtained for the CRF method are significantly higher. The improved results are most likely due to the enhancement of the MATLAB library L-BFGS\footnote{The improvement of the L-BFGS library in 2011 \cite{becker19} has likely resulted in a better learning process of the Conditional Random Field model and, consequently, in an improved algorithm that yields a better performance.} (Limited-memory Broyden–Fletcher–Goldfarb–Shanno \cite{byrd95}).

Recently, Arifoglu et al. \cite{arifoglu17} applied SVMs and different types of Recurrent Neural Networks (RNNs) to the dataset. In their work, only a portion of the data is used for testing, which differs from the approach taken by van Kasteren et al. \cite{kasteren11}, where a full K-Fold cross validation is carried out. The results presented in \cite{kasteren11} are therefore more trustworthy, hence we apply the same technique in this paper.

Singh et al. \cite{singh17} applied an LSTM network to the dataset. Even though the results did not outperform state-of-the-art methods, this work demonstrated that LSTMs are capable of performing well given the temporal dependencies present in this dataset.

Other techniques such as stacked autoencoders \cite{chen18} and modified weighted SVMs \cite{abidine18} have been considered in order to develop a classifier for the dataset. Furthermore, hybrid approaches have also been discussed and applied \cite{guo16, ihianle18, okeyo12, ordonez13, riboni11}.

\subsection{Roadmap}

The rest of the paper is organised as follows. Section~\ref{sec_2} introduces some of the ML models that were used, Section~\ref{sec_3} presents the proposed feature representation and the pre-processing techniques utilised. Section~\ref{experiments_section} demonstrates the effect that the feature representation as well as the combination of different features has on a model's performance. We also show how our best results improve the state-of-the-art. Section~\ref{sec_5} concludes this paper with pointers to future directions.

%% file: preliminary.tex
\section{Preliminary} \label{sec_2}

In this section, we present the task we aim to tackle and provide an overview of some of the ML models applied.

Given a dataset $\{(\mathbf{X}_{t}^{i}, \mathbf{y}_{t})\}$, such that $t=1...T$ and $i=1...N$, where $T$ is the number of data points and $N$ the number of features, the task is to learn a function $f: \mathbb{S}^N \mapsto \{1,...,c\}$, where $S$ is some abstract space and $c$ the number of activities. In this kind of task, both $\langle \mathbf{X}, \mathbf{y} \rangle$ need to be provided in order to perform supervised learning.

For our dataset, $\mathbf{X}$ represents the sensor data and $\mathbf{y}$ the corresponding labels of the activities performed.

\subsection{Probabilistic models}

Na\"ive Bayes, Hidden Markov Model, Hidden Semi-Markov Model and Conditional Random Field constitute the state-of-the-art probabilistic models for this dataset. In the following sections we provide a brief description of those models.

\subsubsection{Na\"ive Bayes}

The Na\"ive Bayes model assumes that data points are independently and identically distributed, which does not account for temporal dependencies or relations between data points with respect to an activity.

Let $\mathbf{X} = (\mathbf{x_{1}},...,\mathbf{x_{T}})$ be a sequence of data points and $\mathbf{y} = (y_{1},...,y_{T})$ the corresponding labels. The joint probability of $\mathbf{y}$ and $\mathbf{X}$ is calculated as follows: 
\[p(\mathbf{y}, \mathbf{X}) = \prod_{t=1}^{T} p(\mathbf{x_{t}}|y_{t})p(y_{t}),\] where $p(\mathbf{x_{t}}|y_{t})$ is decomposed as \[p(\mathbf{x_{t}}|y_{t}) = \prod_{i=1}^{N} p(x_{t}^{i} | y_{t})\] by assuming that the features (e.g. sensors) are conditionally independent given an activity $y_{t}$ (``na\"ive" conditional independence assumption). In other words, sensors $X^i$ and $X^j$ ($i,j \in \{1...N\}$, where $i \neq j$) are conditionally independent given label $y$. This assumption reduces the complexity of the aforementioned classifier however, given that $y$ occurs, knowledge of whether $X^i$ is active provides no information on the likelihood of $X^j$ being active, and vice versa.

\subsubsection{Hidden Markov Model}

The Hidden Markov Model is an extension of Na\"ive Bayes and is capable of modelling temporal dependencies between consecutive time steps. Following the same notation as in the previous section, the model relies on two independence assumptions:
\begin{enumerate}[label=(\roman*)]
    \item $y_{t}$ is only dependent on $y_{t-1}$ (first order Markov assumption);
    \item $\mathbf{x_{t}}$ is only dependent on $y_{t}$ (output independence assumption).
\end{enumerate}

Moreover, the HMM is also a stationary process, which implies that\\$p(y_{t}|y_{t-1}) = p(y_{2}|y_{1}), t \in \{2...T\}$.

The joint probability is calculated as follows:
\[p(\mathbf{y}, \mathbf{X}) = \prod_{t=1}^{T} p(\mathbf{x_{t}}|y_{t})p(y_{t}|y_{t-1})\]

We will use maximum likelihood estimation (MLE) to estimate the parameters $\theta$ which maximises the likelihood of observing $y$ and $X$ given the model $\theta$: $\hat{\theta} = \argmax\limits_{\theta} P(\mathbf{y}, \mathbf{X} | \theta)$.

\subsubsection{Hidden Semi-Markov Model}

A Semi-Markov Model is a generalised Poisson \cite{consul06} process where the holding times need not be independent and identically distributed. Although it is similar to a Markov renewal process \cite{limnios01}, the Hidden Semi-Markov Model (HSMM) \cite{yu15} is a stochastic process where a state has a corresponding length. The length of each state is determined by its duration. Therefore, this is a time-evolving process where the transition between states is made at jump times and dependent upon the corresponding probability distributions.

The main difference between HMMs and HSMMs is the relaxation of the Markov assumption. In particular, HSMMs are able to do this by modelling the duration of a state (e.g. activity). Therefore, a new variable $d_{t}$ is introduced in this model and the joint probability is calculated as follows:
\[p(\mathbf{y}, \mathbf{X}, \mathbf{d}) = \prod_{t=1}^{T} p(\mathbf{x_{t}}|y_{t})p(y_{t}|y_{t-1}, d_{t-1})p(d_{t}|d_{t-1},y_{t}).\]

We use MLE to estimate the parameters $\theta$ which maximises the likelihood of observing $y$, $X$ and $d$ given the model $\theta$: $\hat{\theta} = \argmax\limits_{\theta} P(\mathbf{y}, \mathbf{X}, \mathbf{d} | \theta)$.

\subsubsection{Conditional Random Field}

The Conditional Random Field model, which is the most structurally similar to the HMM model, is called a linear-chain CRF. This model relies on the same independence assumptions as the HMM:
\begin{enumerate}[label=(\roman*)]
    \item $y_{t}$ is only dependent on $y_{t-1}$ (first order Markov assumption);
    \item $\mathbf{x_{t}}$ is only dependent on $y_{t}$ (output independence assumption).
\end{enumerate}

Unlike HSMMs, linear-chain CRF models do not explicitly model the duration of a state. The conditional distribution is calculated using the following expression:
\[p(\mathbf{y} | \mathbf{X}) = \frac{1}{Z(\mathbf{X})} \prod_{t=1}^{T} \exp{\sum_{l=1}^{L} \lambda_{l} f_{l}(y_{t}, y_{t-1}, \mathbf{x_{t}})},\] where $f_{l}(y_{t}, y_{t-1}, \mathbf{x_{t}})$ is a feature function, $\lambda_{l}$ is a weight parameter and $L$ is the number of feature functions. The potential function is the exponential representation of the product of $\lambda_{l} f_{l}(y_{t}, y_{t-1}, \mathbf{x_{t}})$, which can take any positive value, hence why $Z(\mathbf{X})$ is needed as a normalization term.

A CRF is also a stationary process and it uses CMLE (Conditional Maximum Likelihood Estimator), which finds the $\theta$ (CRF parameters) that maximises the conditional likelihood of observing $y$ given the model $\theta$: $\hat{\theta} = \argmax\limits_{\theta} P (\mathbf{y} | \mathbf{X}, \theta)$. Therefore, unlike HMMs which assume that $\mathbf{x_{j}}$ are conditionally independent, CRFs make no assumptions about $p(\mathbf{X})$.

\subsection{Recurrent Neural Network models}

One of the main differences between statistical and neural network models is related to interpretability. Unlike statistical ML models, neural network models do not provide interpretation even though they do provide an effective representation of data properties \cite{karlaftis11}.

In the following sections, three different recurrent neural network models are presented: RNN, LSTM and GRU.

\subsubsection{Recurrent Neural Network}

The RNN model considered is a fully-connected RNN, where the ouput is fed back to the input.
Hence, RNNs contain loops in them which is what allows these type of networks to learn temporal dependencies. 

Let $\mathbf{x} = (x_{1},...,x_{T})$ be an input sequence and $\mathbf{h} = (h_{1},...,h_{T})$ the hidden vector sequence computed by a recurrent neural network. In an RNN, the hidden vector $h^{(t)}$, at time step $t$, is computed as follows:
\[h^{(t)} = \phi(W_{h}h^{(t-1)} + W_{x}x^{(t)}) + b_{h}),\] where $\phi$ is the activation function. The parameters $W$ and $b$ are the weight matrix and bias vector, respectively.

\subsubsection{Long Short-Term Memory Network}

In long-term dependencies, when there is a large time gap between where specific information is stored and where it is needed, RNNs do not perform well; and LSTMs \cite{hochreiter97} are a better and more robust solution. LSTMs are a type of RNNs which are able to detect dependencies across long time windows. 

The LSTM architecture is composed of connected cells and each cell is constituted by three gates: the input ($i^{(t)}$), output ($o^{(t)}$) and forget ($f^{(t)}$) gates, which control the information that is added to or removed from the cell. Moreover, besides having an internal state $c^{(t)}$, a cell also contains a layer which produces the variable $\widetilde{c}^{(t)}$. This variable is representative of the candidate values which may potentially be added to the internal state. 

This type of networks are able to learn the importance of features over time by storing information in the hidden layers. This is done by performing an optimisation of the weights that impacts the information flow. Consequently, LSTMs can lead to a better comprehension of data patterns, which makes them useful to be applied in the field of HAR.

The following equations are used, in an iterative manner, to obtain the scalar value $h^{(t)}$, at time step $t$, of the output vector of the cell. The symbol $\odot$ denotes element-wise multiplication.
\[ i^{(t)} = \sigma(W_{ih}h^{(t-1)} + W_{ix}x^{(t)} + b_{i}) \]
\[ f^{(t)} = \sigma(W_{fh}h^{(t-1)} + W_{fx}x^{(t)} + b_{f}) \]
\[ o^{(t)} = \sigma(W_{oh}h^{(t-1)} + W_{ox}x^{(t)} + b_{o}) \]
\[ \widetilde{c}^{(t)} = \phi(W_{ch}h^{(t-1)} + W_{cx}x^{(t)} + b_{c}) \]
\[ c^{(t)} = i^{(t)} \odot \widetilde{c}^{(t)} + f^{t} \odot c^{(t-1)} \]
\[ h^{(t)} = o^{(t)} \odot \phi(c^{(t)}), \]
where $\sigma$ and $\phi$ are the activation functions.

\subsubsection{Gated Recurrent Unit}

The Gated Recurrent Unit \cite{cho14} is a variation of the LSTM, in which the input and forget gates are combined into one and the cell state and hidden state are the same. Moreover, a new gate called relevance gate is considered in this architecture and it calculates how relevant $c^{(t-1)}$ is to compute $c^{(t)}$. In a GRU, the equations used in order to obtain $h^{(t)}$, at time step $t$, are as follows: 
\[ i^{(t)} = \sigma(W_{ic}c^{(t-1)} + W_{ix}x^{(t)} + b_{i}) \]
\[ r^{(t)} = \sigma(W_{rc}c^{(t-1)} + W_{rx}x^{(t)} + b_{r}) \]
\[ \widetilde{c}^{(t)} = \phi(W_{cc}r^{(t)}c^{(t-1)} + W_{cx}x^{(t)} + b_{c}) \]
\[ c^{(t)} = i^{(t)} \odot \widetilde{c}^{(t)} + (1-i^{(t)}) \odot c^{(t-1)} \]
\[ h^{(t)} = c^{(t)}, \]
where $\sigma$ and $\phi$ are the activation functions.

%% file: propose.tex
\section{Learning from Observation-based Representations} \label{sec_3}

\subsection{The Dataset} \label{exp_thedataset}

The dataset which will be used in the experiments refers to sensor activity in three different houses (A, B and C) \cite{kasteren11}. The data is representative of the activation and deactivation of binary sensors, where a reading is provided every minute for time spans ranging from 14 to 25 days. As a result, in the data there are long stretches where the sensor readings do not change. For example, for houses B and C, on average, the sensors change state only every one and a half hour.

Van Kasteren et al. \cite{kasteren11} used various types of binary sensors (e.g. passive infrared; pressure mats; reed switches), which were placed in three different environments: houses A, B and C. In order to map the observations obtained from these sensors to activities, an annotation system was put in place \cite{kasteren08}.

\begin{figure}
    \caption{Relative frequencies of activities in houses A, B and C}
    \centering
    \subfloat[House A]{\label{fig:housea}\includegraphics[width=0.32\columnwidth]{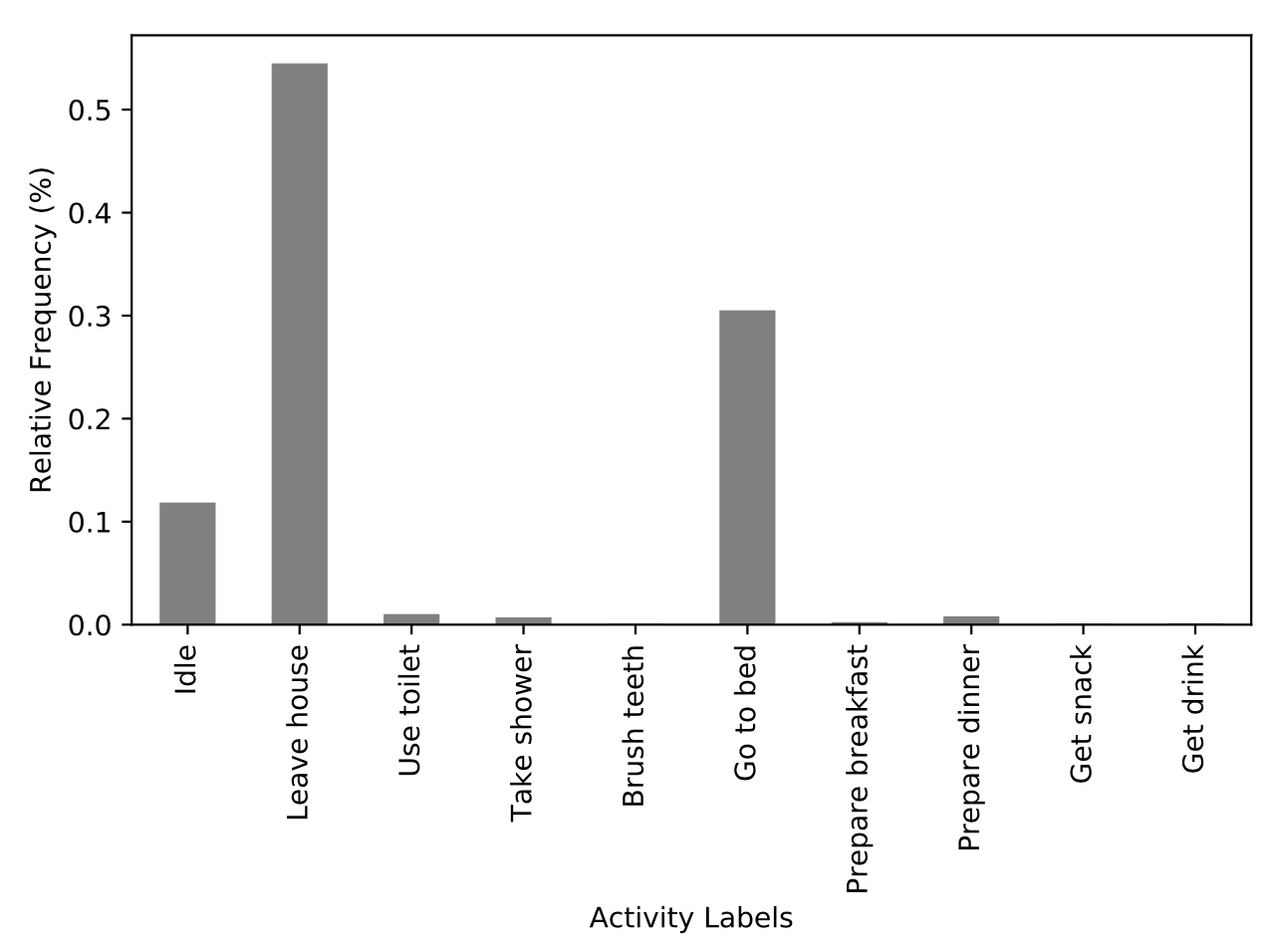}}
    \subfloat[House B]{\label{fig:houseb}\includegraphics[width=0.32\columnwidth]{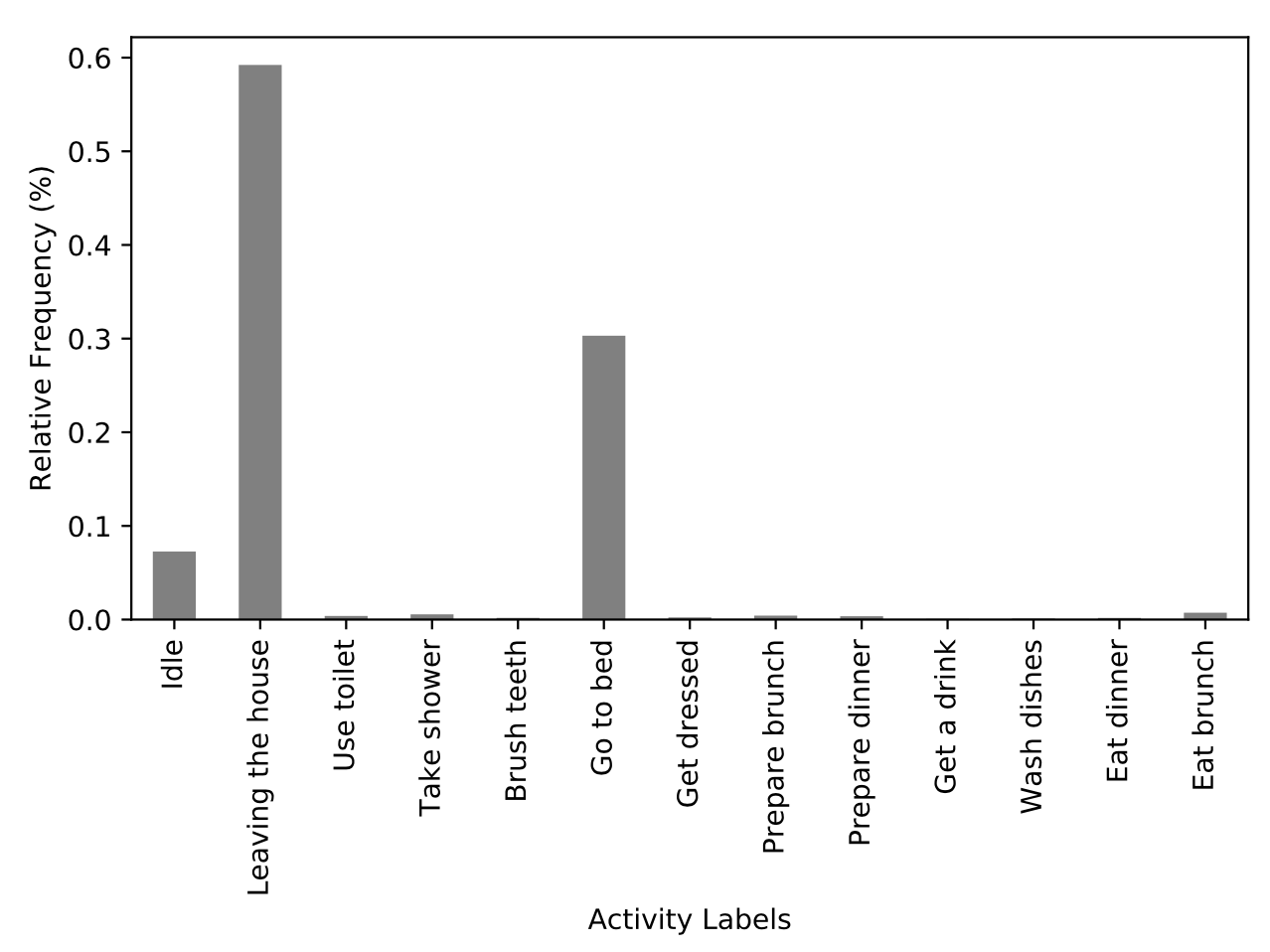}}
    \subfloat[House C]{\label{fig:housec}\includegraphics[width=0.32\columnwidth]{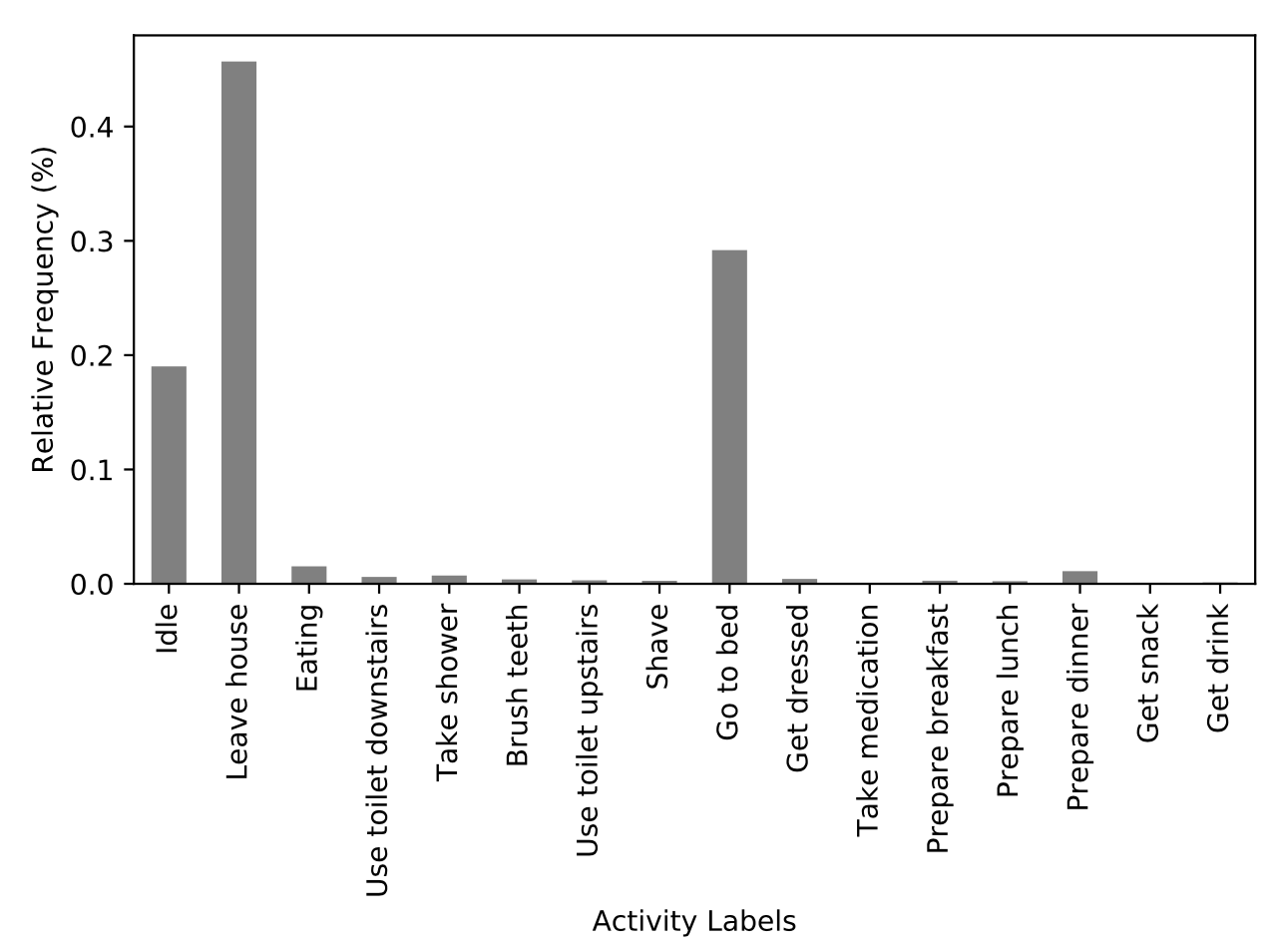}}
    \label{fig:relfreq}
\end{figure}

The relative frequencies of activities in the three different houses are represented in Figure \ref{fig:relfreq}. Table~\ref{info_table} presents some information about this dataset, in particular, the number of sensors placed around the house, the number of activities, the age of the person who inhabited the house and how many days of data we have. In general, the most frequent labels in the three houses are `Idle', `Leave house' and `Go to bed'. A slight higher frequency of label `Idle' is noticeable for house C. On the other hand, the label `Leave the house' acquires a higher frequency in houses A and B.

\subsection{Observation-based Representation}

Since there are long periods of time where the sensors do not change, learning temporal dependencies on this type of data requires a long time history of previous data points, denoted as look-back window. We have observed that there is a gradual increase of training time with higher values for the look-back window. To overcome this, we propose a new representation for sensor data called observation-based (OB) representation, which combines consecutive data points with the same sensor readings into one data point. Hence, data points are merged if sensor readings remain unchanged.

\begin{table}
    \caption{Details about the dataset}
    \label{info_table}
    \begin{tabularx}{\columnwidth}{*5{C}}
        \toprule
        House & Sensors & Activities & Age & Duration (days)\\
        \midrule
        A & 14 & 10 & 26 & 25\\ 
        B & 23 & 13 & 28 & 14\\
        C & 21 & 16 & 57 & 19\\
        \bottomrule
    \end{tabularx}
\end{table}

Furthermore, three different feature representations were considered in \cite{kasteren11}: raw, changepoint and last-fired. These were initially introduced in \cite{kasteren08} and are a way of comparing how the data is given as an input and the impact that it has in the overall recognition performance. In the raw representation, the sensor takes value 1 when it is activated and 0 otherwise; with the changepoint representation, the sensor takes value 1 when it changes state and 0 otherwise; the last fired representation makes the last sensor that changed state to take value 1 until another sensor changes its state.

In comparison, our proposed representation is more expressive than the changepoint and last-fired representations, because it yields information about the current and/or most recent sensors that have changed its value, without having to provide a large number for the look-back window. The disadvantage of having a large number for the look-back window is that it may affect the classification of other activities which do not require all the information provided by the data that is fed into the network.

\subsection{Time Related Features} \label{preprocessing}

When computing the OB representation, the variable $\Delta t$ is obtained by calculating how long the sensor readings remain unchanged. Since the dataset provides sensor readings in 60 second intervals, $\Delta t$ indicates the duration (in minutes) of no change for a sensor reading. We study the effect of using this variable as well as the $hour$ variable, which represents the hour of a sensor reading. By incorporating the latter, the information provided can be useful for classification purposes.

\begin{figure}
    \caption{Frequencies of the $\Delta$t variable for houses A, B and C}
    \centering
    \subfloat[House A]{\label{fig:housea_deltat}\includegraphics[width=0.32\columnwidth]{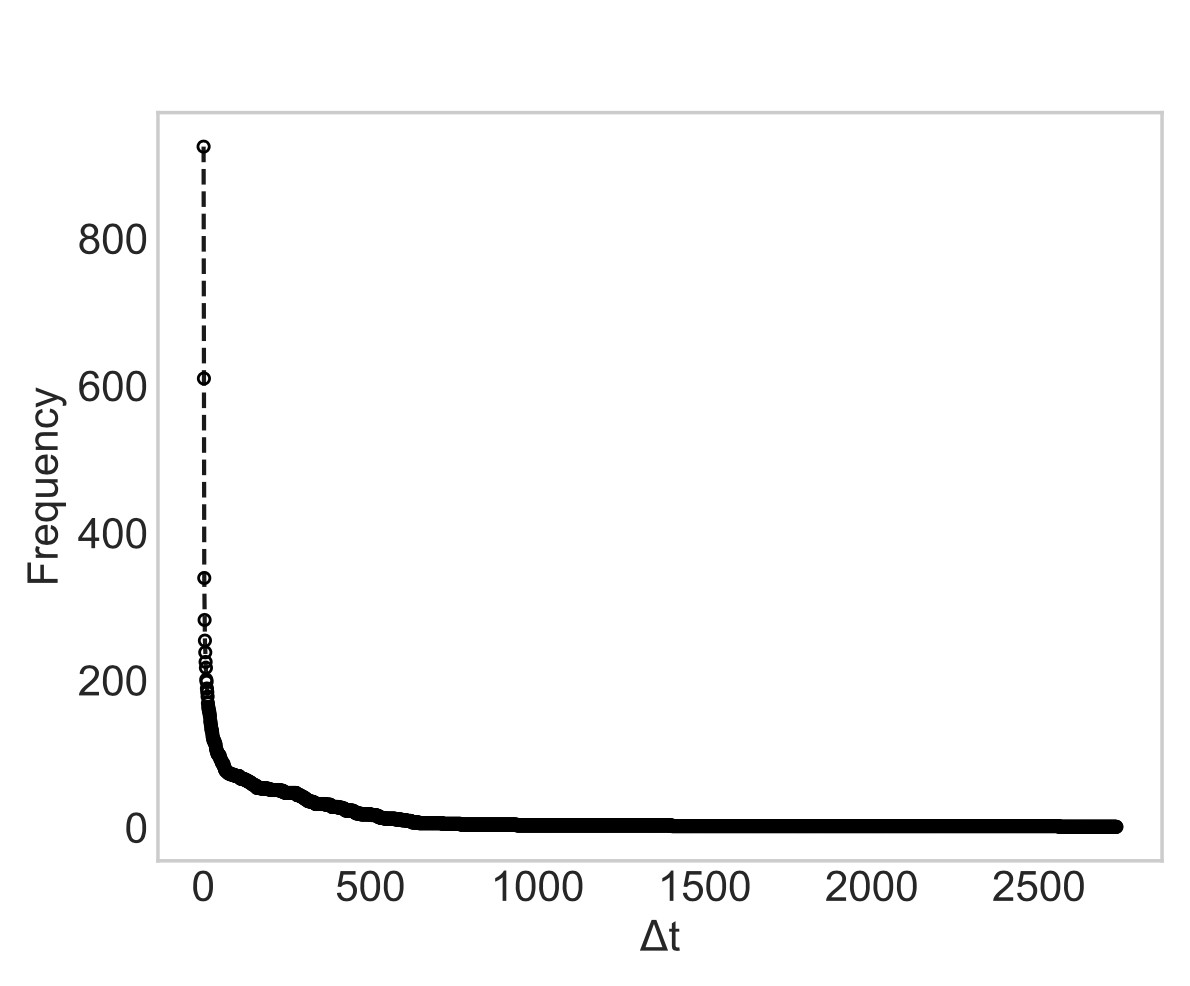}}
    \subfloat[House B]{\label{fig:houseb_deltat}\includegraphics[width=0.32\columnwidth]{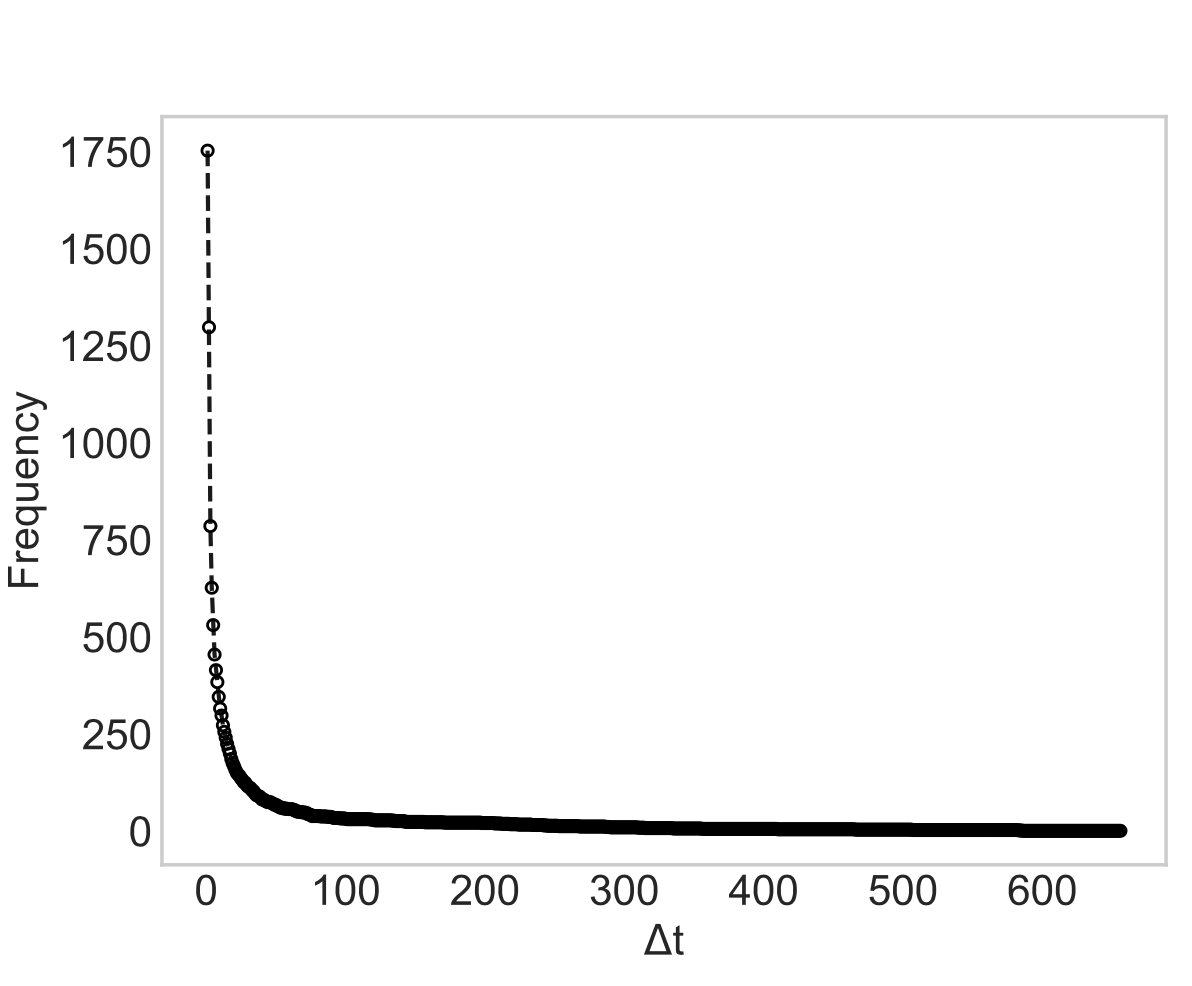}}
    \subfloat[House C]{\label{fig:housec_deltat}\includegraphics[width=0.32\columnwidth]{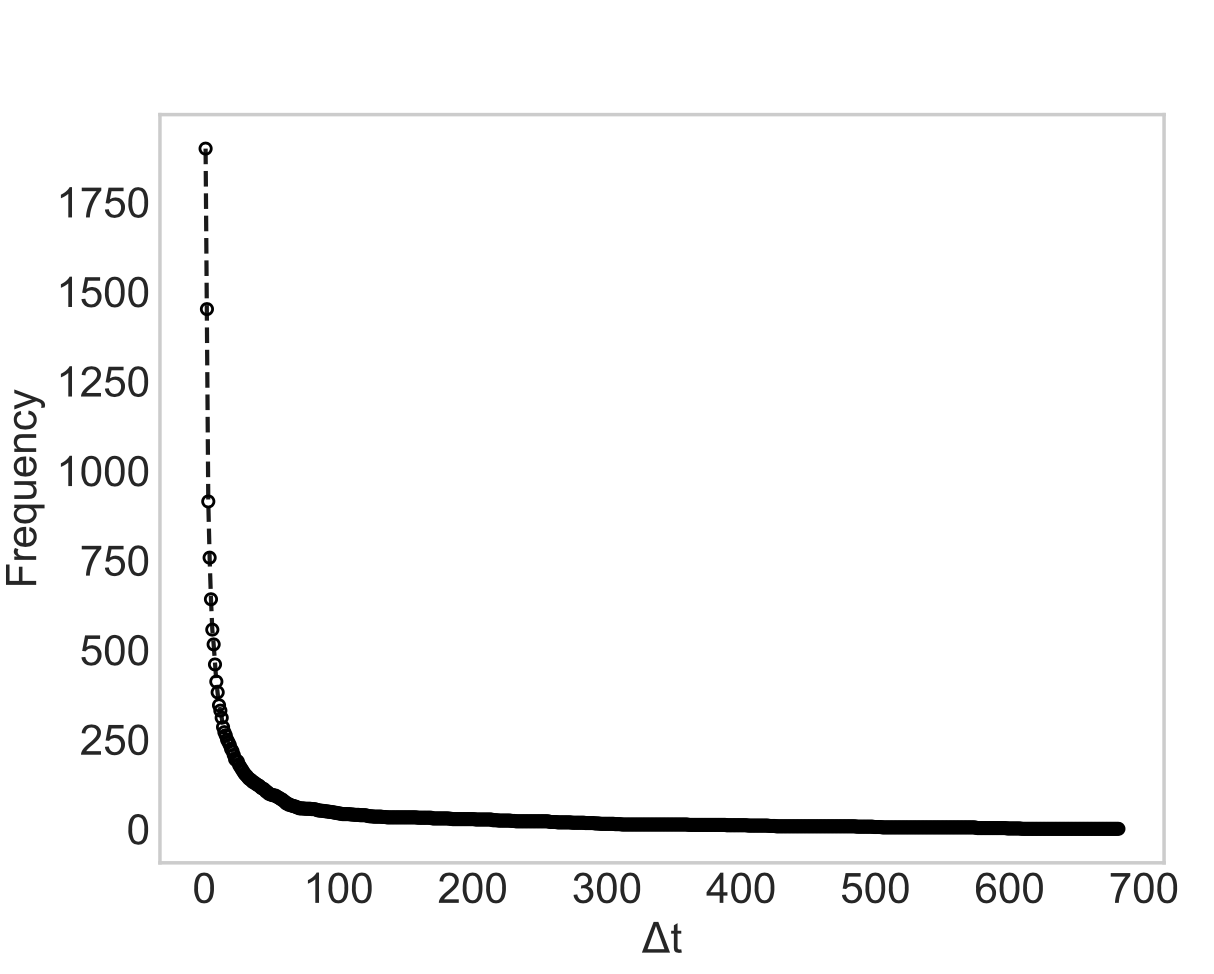}}
    \label{fig:deltat_distr}
\end{figure}

The frequency of each possible value for variable $\Delta t$ in house A is presented in Figure~\ref{fig:housea_deltat} and we observe a similar distribution for houses B and C (Figures~\ref{fig:houseb_deltat} and \ref{fig:housec_deltat}). For house A, this variable can take values from $1$ to $2732$. In order to keep the number of features small, we further discretize $\Delta t$ into coarser bins. Hence, each bin will essentially represent an interval. Based on the relative frequency, we considered two different ways of splitting this variable into intervals: one results in a total of 48 intervals (48i) and the other one in a total of 7 (7i). The difference between the two lies on the importance of categorising smaller durations. Let $t_{i} \in \Delta t$. In 48i, the following cases were considered:
\begin{enumerate}
    \item Each $t_{i}$ is uniquely encoded if $t_{i} <= 30$;
    \item One encoding representation for each of the following intervals:
\end{enumerate} 
\vspace{-0.05\textwidth}
\begin{center}
    \footnotesize
    \begin{tabular}{p{0.33\linewidth}p{0.33\linewidth}p{0.33\linewidth}}
        \multicolumn{1}{l}{(a) $30 < t_{i} <= 40$} & \multicolumn{1}{l}{(g) $120 < t_{i} <= 150$} & \multicolumn{1}{l}{(m) $300 < t_{i} <= 360$}\\
        (b) $40 < t_{i} <= 50$ & (h) $150 < t_{i} <= 180$ & (n) $360 < t_{i} <= 420$\\
        (c) $50 < t_{i} <= 60$ & (i) $180 < t_{i} <= 210$ & (o) $420 < t_{i} <= 480$\\
        (d) $60 < t_{i} <= 80$ & (j) $210 < t_{i} <= 240$ & (p) $480 < t_{i} <= 540$\\
        (e) $80 < t_{i} <= 100$ & (k) $240 < t_{i} <= 270$ & (q) $540 < t_{i} <= 600$\\
        (f) $100 < t_{i} <= 120$ & (l) $270 < t_{i} <= 300$ & (r) $t_{i} > 600$\\
  \end{tabular}
\end{center}

On the other hand, for 7i, each of the intervals below were uniquely encoded:
\vspace{-0.03\textwidth}
\begin{center}
    \footnotesize
    \begin{tabular}{p{0.33\linewidth}p{0.33\linewidth}p{0.33\linewidth}}
        \multicolumn{1}{l}{(a) $t_{i} <= 5$} & \multicolumn{1}{l}{(d) $60 < t_{i} <= 120$} & \multicolumn{1}{l}{(f) $150 < t_{i} <= 660$}\\
        (b) $5 < t_{i} <= 30$ & (e) $120 < t_{i} <= 150$ & (g) $t_{i} > 660$\\
        (c) $30 < t_{i} <= 60$ & & \\
  \end{tabular}
\end{center}

We then encode each interval considering two different encoding processes: one-hot and unary-based encodings. Regarding the one-hot encoding process, it will generate a squared matrix, where the number of rows is the same as the number of values. Therefore, it creates new binary columns, indicating the presence of each possible value. As for the unary-based encoding, this process also creates a squared matrix which has the same dimension as the matrix generated in the previous encoding process. The main difference between these two encoding processes lies on the interpretation of the binary columns. In the one-hot encoding, the binary columns indicate the presence of each possible value, therefore only one component in each column will take value one. On the other hand, for the unary-based coding, the binary columns indicate the presence of values that are less than or equal to each possible value; hence, without loss of generality, supposing the values are in ascending order, all the elements of the lower triangle of the matrix will take value one.

In regard to the $hour$ variable, which can take values from $0$ to $23$, we also encode this variable using the two processes aforementioned (one-hot and unary-based encodings) but we consider each number a category so, for this particular variable, we will have exactly 24 values. Hence, each category will be representative of the hour of the sensor reading. The reason why we encode this variable such that the values of the features are in the same range as the other features is because this makes training faster and reduces the chances of getting stuck in local optima.

%% file: experiment.tex
\section{Experiments} \label{experiments_section} 

In the following experiments, an OB representation of the dataset is used in order to compare and evaluate against other feature representations. The OB representation is obtained by directly collecting information from the sensors, which corresponds to the data in its raw representation format. As demonstrated in Table \ref{sotares_table} (Appendix \ref{appendix}), the raw representation gives the worst results, irrespective of the algorithm. Hence, a good performance by both generative and discriminant algorithms is always dependent on considering a changepoint or last-fired representation. The OB representation provides a generalisation of the changepoint and last-fired representations. Some further discussion of the proposed method and analysis of the results is presented in the following sections.

These experiments were run using a K-Fold cross validation approach, where we cycle through each one of the days using it for testing and the data corresponding to the remaining days is used for training. This is consistent with the technique applied by van Kasteren et al. \cite{kasteren11}. The mean per class accuracy as well as the overall accuracy are presented as evaluation metrics; and the accuracy for each class is also calculated.

In regard to the neural network models - RNN, LSTM, GRU, MLP and LSTM with a CRF layer (LSTMCRF) - we considered the following set of hyper-parameters: $128$ for the number of units, a learning rate of $0.0001$, $100$ for the number of epochs and a batch size of $512$.

The optimisation algorithms that were used in order to minimise the error rates of the ML models were the Root Mean Square Propagation for training the RNN, LSTM and GRU models, and the Adaptive Moment Estimation optimiser was used for training the MLP and LSTMCRF models.

These parameters were selected after analysing the training losses and accuracies of the models applied by taking into consideration their performances across different sets of hyper-parameters. We also aimed at making a fair comparison among these methods and selected the same set of hyper-parameters for the NN-based models. Moreover, these particular parameters have shown to work relatively well for these methods irrespective of the feature representation.

\subsection{Evaluation metrics}

We will be using the mean per class accuracy and the \emph{accuracy} as evaluation metrics for our experiments. The latter can be defined as follows. Let $pred$ and $true$ be the $N$-dimensional arrays which contain the model's predictions and the true labels of each data point, respectively. Then, the accuracy is the percentage of correctly predicted activities, i.e.:
\[
\mathsf{accuracy} = \frac{|\{i \in \{1,...,N\} \mid \mathsf{pred}(i) = \mathsf{true}(i)\}|}{N}.
\]

Given the imbalance of the dataset, a classifier would not be properly evaluated if accuracy was the only metric utilised to assess its performance. Therefore, the accuracy for each class is also presented in order to analyse whether the models are being able to accurately classify not only highly frequent classes but also infrequent ones. 

Formally, the accuracy of a class c is given by 
\[
\mathsf{accuracy_c} = \frac{| \{i \in \{1,...,N_{c}\} \mid \mathsf{pred_{c}}(i) = \mathsf{true_{c}}(i) \} |}{| \{ i \in \{1,...,N\} \mid \mathsf{true}(i) = c \} |},
\]
where $pred_{c}$ and $true_{c}$ are the $N_{c}$-dimensional arrays which contain the model's predictions and the true labels of each data point belonging to a class $c$, respectively.

Lastly, we define the mean per class accuracy as follows. Let $c \in \{1,...,C\}$, where $C$ is the number of activities in a dataset. Then, the mean per class accuracy is calculated according to the following expression:
\[
\mathsf{mean\_per\_class\_accuracy} = \frac{1}{c} \sum\limits_{c=1}^{c} \mathsf{accuracy_c}.
\]

The best values for the mean per class accuracy, overall accuracy and per-class accuracies are highlighted in bold.

\subsection{Effect of the Feature Representation} \label{rep_effect}

All the experiments presented in this section do not take into consideration the features $hour$ (\textit{NoToD}, where \textit{ToD} stands for \textit{Time of Day}) nor the $\Delta t$ (\textit{NoDeltaT}), i.e. the features $hour$ and $\Delta t$ were not added to the dataset.

\subsubsection{Raw Feature Representation}

In the following tables (Tables \ref{labelacc_houseA1}, \ref{labelacc_houseB1} and \ref{labelacc_houseC1}), we evaluate 8 different methods using a raw feature representation: NB, HMM, HSMM, CRF, LSTM, GRU, RNN and LSTMCRF. We considered a look-back window of 1 and this serves as a baseline for the experiments run in the next subsections. In particular, the results provided by the methods NB, HMM, HSMM and CRF were obtained by reproducing the experiments done in \cite{kasteren11}. The CRF model outperformed the other models for houses A and C. Specifically, the accuracy(mean per class accuracy) achieved for house A was 91.85$\pm$7.80(59.13 $\pm$ 15.66) and for house C 73.83$\pm$22.39(32.03$\pm$20.23). For house B, the RNN model achieved the best overall accuracy (87.16$\pm$11.12) in comparison to the other models; however, the CRF model provided the best value for the mean per-class accuracy (47.64$\pm$13.17).
\begin{table}
    \caption{Accuracy per label (Raw feature representation) - House A (Look-back window: 1)}
    \begin{adjustbox}{width=\columnwidth,center}
        \begin{tabularx}{1.6\textwidth}{c *{9}{C}}
            \toprule
            Label & NB & HMM & HSMM & CRF & LSTM & GRU & RNN & LSTMCRF\\ 
            \midrule
            `Idle' & 10.65 & 49.01 & 55.03 & \textbf{69.53} & 14.5 & 17.76 & 13.1 & 17.85\\
            `Leave house' & 91.29 & 42.94 & 41.99 & \textbf{98.73} & 98.1 & 98.1 & 98.1 & 98.1\\
            `Use toilet' & 44.11 & 70.68 & \textbf{73.97} & 35.62 & 43.84 & 42.74 & 42.74 & 43.29\\
            `Take shower' & 0.0 & 0.0 & 0.0 & 0.0 & 0.0 & 0.0 & 0.0 & 0.0\\
            `Brush teeth' & 0.0 & \textbf{15.62} & 12.5 & 0.0 & 0.0 & 0.0 & 0.0 & 0.0\\
            `Go to bed' & 83.18 & 93.08 & 93.08 & 94.87 & 95.25 & \textbf{95.25} & \textbf{95.25} & \textbf{95.25}\\
            `Prepare breakfast' & 45.98 & 48.28 & 51.72 & \textbf{63.22} & 48.28 & 45.98 & 40.23 & 48.28\\
            `Prepare dinner' & 10.8  & 29.97 & 37.28 & \textbf{50.87} & 9.76 & 11.15 & 11.5 & 10.8\\
            `Get snack' & 14.29 & 28.57 & \textbf{54.76} & 40.48 & 0.0 & 0.0 & 7.14 & 0.0\\
            `Get drink' & 34.69 & 40.82 & 57.14 & \textbf{71.43} & 32.65 & 51.02 & 53.06 & 40.82\\
            \midrule
            Mean per class accuracy & $42.46$ & $45.18$ & $47.97$ & $\mathbf{59.13}$ & $47.63$ & $48.62$ & $48.22$ & $48.55$\\
            Standard deviation & $16.64$ & $19.11$ & $19.27$ & $15.66$ & $17.19$ & $16.95$ & $16.92$ & $16.96$\\
            \midrule
            Overall accuracy & $77.13$ & $59.08$ & $58.46$ & $\mathbf{91.85}$ & $84.94$ & $85.34$ & $84.80$ & $85.35$\\
            Standard deviation & $20.80$ & $28.64$ & $29.25$ & $7.80$ & $10.80$ & $10.88$ & $11.14$ & $10.92$\\
            \bottomrule
        \end{tabularx}
    \end{adjustbox}
    \label{labelacc_houseA1}
\end{table}
\begin{table}
    \caption{Accuracy per label (Raw feature representation) - House B (Look-back window: 1)}
    \begin{adjustbox}{width=\columnwidth,center}
        \begin{tabularx}{1.6\textwidth}{c *{9}{C}}
            \toprule
            Label & NB & HMM & HSMM & CRF & LSTM & GRU & RNN & LSTMCRF\\ 
            \midrule
            `Idle' & 33.79 & 28.62 & 37.38 & \textbf{55.52} & 31.45 & 32.69 & 34.69 & 28.83\\
            `Leaving the house' & 87.48 & 59.29 & 59.27 & 79.19 & \textbf{96.97} & 96.34 & 96.45 & 92.51\\
            `Use toilet' & 19.48 & \textbf{40.26} & 38.96 & 20.78 & 0.0 & 1.3 & 12.99 & 0.0\\
            `Take shower' & 13.51 & \textbf{69.37} & 65.77 & 55.86 & 0.0 & 9.01 & 20.72 & 0.0\\
            `Brush teeth' & 0.0 & \textbf{30.56} & 27.78 & 5.56 & 0.0 & 0.0 & 0.0 & 0.0\\
            `Go to bed' & 84.24 & 80.9 & 80.93 & 83.87 & 87.34 & 87.34 & \textbf{89.16} & 86.27\\
            `Get dressed' & 47.83 & \textbf{65.22} & 63.04 & 39.13 & 0.0 & 6.52 & 39.13 & 4.35\\
            `Prepare brunch' & 15.48  & 30.95 & 30.95 & \textbf{51.19} & 14.29 & 13.1 & 10.71 & 8.33\\
            `Prepare dinner' & 15.49 & 36.62 & 36.62 & \textbf{53.52} & 5.63 & 18.31 & 25.35 & 4.23\\
            `Get a drink' & 0.0 & \textbf{35.71} & \textbf{35.71} & 14.29 & 0.0 & 0.0 & 0.0 & 0.0\\ 
            `Wash dishes' & 0.0 & \textbf{4.76} & 0.0 & 0.0 & 0.0 & 0.0 & 0.0 & 0.0\\ 
            `Eat dinner' & 0.0 & 0.0 & 0.0 & \textbf{5.71} & 0.0 & 0.0 & 0.0 & 0.0\\ 
            `Eat brunch' & 0.0 & 18.49 & 19.86 & \textbf{25.34} & 0.0 & 0.0 & 0.0 & 0.0\\
            \midrule
            Mean per class accuracy & $32.65$ & $43.77$ & $44.76$ & $\mathbf{47.64}$	& $30.3$	& $32.26$	& $37.84$	& $27.93$\\
            Standard deviation & $8.26$ & $12.84$ & $13.62$ & $13.17$	& $13.42$	& $12.31$	& $11.14$	& $12.68$\\
            \midrule
            Overall accuracy & $80.37$ & $63.34$ & $63.78$ & $79.59$ & $86.39$ & $86.21$ & $\mathbf{87.16}$ & $83.24$\\
            Standard deviation & $17.97$ & $24.49$ & $24.27$ & $23.84$ & $11.41$ & $11.12$ & $11.12$ & $18.85$\\
            \bottomrule
        \end{tabularx}
    \end{adjustbox}
    \label{labelacc_houseB1}
\end{table}
\begin{table}
    \caption{Accuracy per label (Raw feature representation) - House C (Look-back window: 1)}
    \begin{adjustbox}{width=\columnwidth,center}
        \begin{tabularx}{1.6\textwidth}{c *{9}{C}}
            \toprule
            Label & NB & HMM & HSMM & CRF & LSTM & GRU & RNN & LSTMCRF\\ 
            \midrule
            `Idle' & 37.04 & 13.23 & 24.25 & \textbf{52.3} & 26.54 & 25.5 & 25.12 & 25.66\\
            `Leave house' & 54.32 & 23.98 & 25.86 & \textbf{71.64} & 49.5 & 41.81 & 50.03 & 50.05\\
            `Eating' & 1.0 & 9.23 & 15.21 & \textbf{23.94} & 1.5 & 2.24 & 1.0 & 1.5\\
            `Use toilet downstairs' & 0.0 & 0.63 & \textbf{1.27} & 0.0 & 0.0 & 0.0 & 0.0 & 0.0\\
            `Take shower' & 1.58 & \textbf{11.58} & 11.05 & 9.47 & 0.0 & 0.0 & 0.53 & 0.0\\
            `Brush teeth' & 8.91 & 23.76 & \textbf{25.74} & 14.85 & 0.99 & 1.98 & 4.95 & 0.0\\
            `Use toilet upstairs' & 8.75 & 21.25 & \textbf{26.25} & 20.0 & 2.5 & 5.0 & 7.5 & 1.25\\
            `Shave' & 0.0 & 24.64 & 21.74 & \textbf{46.38} & 0.0 & 0.0 & 0.0 & 0.0\\
            `Go to bed' & 51.28 & 45.83 & 52.05 & \textbf{67.34} & 61.49 & 60.16 & 59.56 & 62.54\\
            `Get dressed' & 8.93 & \textbf{32.14} & 31.25 & 8.04 & 5.36 & 5.36 & 4.46 & 3.57\\
            `Take medication' & 0.0 & 0.0 & 0.0 & 0.0 & 0.0 & 0.0 & 0.0 & 0.0\\
            `Prepare breakfast' & 0.0  & 5.63 & 5.63 & \textbf{16.9} & 0.0 & 0.0 & 0.0 & 0.0\\
            `Prepare lunch' & 0.0  & 15.0 & \textbf{23.33} & 0.0 & 0.0 & 0.0 & 0.0 & 0.0\\
            `Prepare dinner' & 4.14 & 11.03 & 9.66 & \textbf{45.86} & 6.21 & 5.86 & 6.21 & 0.0\\
            `Get snack' & 0.0 & 8.33 & \textbf{16.67} & 0.0 & 0.0 & 0.0 & 0.0 & 0.0\\
            `Get drink' & 0.0 & 0.0 & 0.0 & \textbf{16.13} & 0.0 & 0.0 & 0.0 & 0.0\\
            \midrule
            Mean per class accuracy & $16.85$	& $17.24$	& $20.61$	& $\mathbf{32.03}$ & $15.06$ & $14.94$ & $15.26$ & $15.19$\\
            Standard deviation & $7.56$	& $9.49$	& $11.00$	& $20.23$ & $7.77$ & $8.48$ & $7.23$ & $7.86$\\
            \midrule
            Overall accuracy & $46.49$ & $26.48$ & $31.40$ & $\mathbf{73.83}$ & $45.28$ & $41.36$ & $44.73$ & $45.67$\\
            Standard deviation & $22.56$ & $22.75$ & $24.62$ & $22.39$ & $23.53$ & $24.31$ & $22.75$ & $23.23$\\
            \bottomrule
        \end{tabularx}
    \end{adjustbox}
    \label{labelacc_houseC1}
\end{table}

\subsubsection{Observation-based representation with RNN-based methods}

We have also applied the LSTM, GRU, RNN and LSTMCRF methods to the raw and OB feature representations. We considered the following values for the look-back window: 2, 5 and 10. In Tables~\ref{labelacc_houseA_RawvsOB_LSTMGRU}, \ref{labelacc_houseA_RawvsOB_RNNLSTMCRF}, \ref{labelacc_houseB_RawvsOB_LSTMGRU}, \ref{labelacc_houseB_RawvsOB_RNNLSTMCRF}, \ref{labelacc_houseC_RawvsOB_LSTMGRU} and \ref{labelacc_houseC_RawvsOB_RNNLSTMCRF}, we present the results achieved for houses A, B and C across different look-back window values: 2, 5 and 10.

\begin{table}
    \caption{Accuracy per label using LSTM and GRU (Raw vs OB feature representations) - House A (Look-back window: 2, 5 and 10)}
    \begin{adjustbox}{width=\columnwidth,center}
        \begin{tabularx}{1.6\textwidth}{c *{13}{C}}
            \toprule
            \multicolumn{1}{c}{} & \multicolumn{6}{c}{LSTM} & \multicolumn{6}{c}{GRU}\\
            \cmidrule(lr){2-7} \cmidrule(l){8-13}
            \multicolumn{1}{c}{Label} & \multicolumn{2}{c}{2} & \multicolumn{2}{c}{5} & \multicolumn{2}{c}{10} & \multicolumn{2}{c}{2} & \multicolumn{2}{c}{5} & \multicolumn{2}{c}{10}\\
            \cmidrule(lr){2-3} \cmidrule(l){4-5} \cmidrule(l){6-7} \cmidrule(l){8-9} \cmidrule(l){10-11} \cmidrule(l){12-13}
             & Raw & OB & Raw & OB & Raw & OB & Raw & OB & Raw & OB & Raw & OB\\
            \midrule
            `Idle' & 25.27 & 87.69 & 31.12 & 81.29 & 39.07 & 68.29 & 26.13 & 86.26 & 30.57 & 81.51 & 37.9 & 68.6\\
            `Leave house' & 96.46 & 99.86 & 96.37 & 98.9 & 96.18 & 87.78 & 96.46 & 99.88 & 96.38 & 98.96 & 96.24 & 99.75\\
            `Use toilet' & 42.19 & 62.47 & 43.01 & 57.53 & 40.55 & 18.36 & 45.21 & 64.11 & 49.04 & 59.73 & 47.12 & 57.26\\
            `Take shower' & 0.0 & 0.0 & 0.0 & 5.58 & 5.58 & 10.76 & 0.0 & 0.0 & 0.0 & \textbf{25.5} & 0.0 & 23.11\\
            `Brush teeth' & 0.0 & 0.0 & 0.0 & 0.0 & 0.0 & 0.0 & 0.0 & 0.0 & 0.0 & 0.0 & 0.0 & 0.0\\
            `Go to bed' & 95.23 & \textbf{97.76} & 95.18 & 95.16 & 95.29 & 90.28 & 95.24 & \textbf{97.76} & 95.2 & 95.12 & 95.36 & 93.35\\
            `Prepare breakfast' & 55.17 & 56.32 & 52.87 & 50.57 & 45.98 & 18.39 & 57.47 & 56.32 & 60.92 & 54.02 & \textbf{65.52} & 58.62\\
            `Prepare dinner' & 12.54 & 9.76 & 9.06 & 11.85 & 11.5 & 10.45 & 13.59 & 14.29 & 15.68 & 22.3 & 13.59 & 21.95\\
            `Get snack' & 0.0 & 0.0 & 0.0 & 7.14 & 0.0 & 0.0 & 0.0 & 0.0 & 0.0 & \textbf{14.29} & 0.0 & 9.52\\
            `Get drink' & 28.57 & 34.69 & 24.49 & 4.08 & 12.24 & 0.0 & \textbf{44.9} & 42.86 & 36.73 & 30.61 & 42.86 & 26.53\\
            \midrule
            Mean per class accuracy & $48.70$	& $62.03$	& $49.22$	& $58.63$ & $49.44$ & $45.18$ & $50.90$ & $62.79$ & $52.59$ & $\mathbf{63.09} $ & $53.98$ & $61.91$\\
            Standard deviation & $17.15$	& $13.88$	& $17.13$	& $16.0$ & $16.56$ & $19.04$ & $16.61$ & $13.71$ & $16.37$ & $15.71$ & $16.53$ & $15.97$\\
            \midrule
            Overall accuracy & $85.33$ & $\mathbf{95.54}$ & $85.91$ & $93.46$ & $86.77$ & $83.95$ & $85.50$ & $95.45$ & $86.01$ & $93.81$ & $86.82$ & $92.13$\\
            Standard deviation & $10.50$ & $3.25$ & $10.31$ & $6.08$ & $9.90$ & $12.96$ & $10.43$ & $3.37$ & $10.26$ & $5.48$ & $9.89$ & $7.06$\\
            \bottomrule
        \end{tabularx}
    \end{adjustbox}
    \label{labelacc_houseA_RawvsOB_LSTMGRU}
\end{table}

\begin{table}
    \caption{Accuracy per label using RNN and LSTMCRF (Raw vs OB feature representations) - House A (Look-back window: 2, 5 and 10)}
    \begin{adjustbox}{width=\columnwidth,center}
        \begin{tabularx}{1.6\textwidth}{c *{13}{C}}
            \toprule
            \multicolumn{1}{c}{} & \multicolumn{6}{c}{RNN} & \multicolumn{6}{c}{LSTMCRF}\\
            \cmidrule(lr){2-7} \cmidrule(l){8-13}
            \multicolumn{1}{c}{Label} & \multicolumn{2}{c}{2} & \multicolumn{2}{c}{5} & \multicolumn{2}{c}{10} & \multicolumn{2}{c}{2} & \multicolumn{2}{c}{5} & \multicolumn{2}{c}{10}\\
            \cmidrule(lr){2-3} \cmidrule(l){4-5} \cmidrule(l){6-7} \cmidrule(l){8-9} \cmidrule(l){10-11} \cmidrule(l){12-13}
             & Raw & OB & Raw & OB & Raw & OB & Raw & OB & Raw & OB & Raw & OB\\
            \midrule
            `Idle' & 25.94 & 86.0 & 30.55 & 66.18 & 39.19 & 55.41 & 25.91 & \textbf{89.09} & 30.4 & 77.2 & 40.58 & 69.08\\
            `Leave house' & 96.45 & 99.88 & 96.37 & 98.89 & 96.17 & 97.62 & 96.47 & \textbf{99.89} & 96.36 & 94.52 & 96.12 & 48.96\\
            `Use toilet' & 44.66 & \textbf{67.67} & 52.88 & 63.84 & 47.95 & 55.62 & 45.75 & 55.89 & 54.25 & 1.64 & 47.12 & 0.82\\
            `Take shower' & 0.0 & 0.0 & 0.4 & 22.31 & 7.57 & 22.31 & 0.0 & 0.0 & 0.0 & 1.2 & 0.0 & 4.78\\
            `Brush teeth' & 0.0 & 0.0 & 0.0 & 0.0 & 0.0 & 0.0 & 0.0 & 0.0 & 0.0 & 0.0 & 0.0 & 0.0\\
            `Go to bed' & 95.21 & \textbf{97.76} & 95.21 & 94.33 & 95.44 & 91.89 & 95.17 & \textbf{97.76} & 95.43 & 85.66 & 95.5 & 74.45\\
            `Prepare breakfast' & 49.43 & 54.02 & 52.87 & 48.28 & 55.17 & 44.83 & 63.22 & 48.28 & 59.77 & 1.15 & 48.28 & 0.0\\
            `Prepare dinner' & 9.76 & 15.33 & 12.54 & \textbf{26.13} & 13.24 & 23.69 & 12.54 & 5.92 & 12.2 & 1.05 & 5.23 & 12.54\\
            `Get snack' & 0.0 & 4.76 & 0.0 & 4.76 & 2.38 & 7.14 & 0.0 & 0.0 & 2.38 & 0.0 & 7.14 & 0.0\\
            `Get drink' & 34.69 & 34.69 & 32.65 & 28.57 & 38.78 & 6.12 & 34.69 & 0.0 & 28.57 & 0.0 & 28.57 & 0.0\\
            \midrule
            Mean per class accuracy & $48.68$	& $62.58$	& $51.22$	& $60.70$	& $53.68$ & $55.54$ & $50.80$ & $57.37$ & $52.48$ & $41.38$ & $51.43$	& $32.73$\\
            Standard deviation & $17.42$	& $13.81$	& $16.93$	& $17.50$	& $16.42$ & $17.87$ & $16.80$ & $14.86$ & $16.77$ & $19.40$ & $17.10$	& $22.64$\\
            \midrule
            Overall accuracy & $85.39$ & $95.46$ & $86.0$ & $91.77$ & $87.0$ & $88.86$ & $85.46$ & $95.56$ & $86.08$ & $86.96$ & $87.04$ & $57.70$\\
            Standard deviation & $10.49$ & $3.44$ & $10.25$ & $7.91$ & $9.97$ & $9.0$ & $10.45$ & $3.36$ & $10.23$ & $12.38$ & $9.93$ & $26.24$\\
            \bottomrule
        \end{tabularx}
    \end{adjustbox}
    \label{labelacc_houseA_RawvsOB_RNNLSTMCRF}
\end{table}

In regard to house A, we observe that, for all methods, this dataset does not require a large value for the look-back window in order to be able to accurately classify highly frequent labels. LSTM is the method which provides the highest accuracy considering a look-back window of $2$. Also, considering the mean per class accuracy, GRUs are able to perform better than any of the other RNN-based methods. We observe that the optimal value for the look-back window here was $5$, which only differs 0.3 percent points from the result obtained for the same method with a look-back window of $2$; therefore, since the difference between the mean per-class accuracies is not significant, a small look-back window provides enough knowledge in order to achieve a good performance in this classification task.

\begin{table}
    \caption{Accuracy per label using LSTM and GRU (Raw vs OB feature representations) - House B (Look-back window: 2, 5 and 10)}
    \begin{adjustbox}{width=\columnwidth,center}
        \begin{tabularx}{1.6\textwidth}{c *{13}{C}}
            \toprule
            \multicolumn{1}{c}{} & \multicolumn{6}{c}{LSTM} & \multicolumn{6}{c}{GRU}\\
            \cmidrule(lr){2-7} \cmidrule(l){8-13}
            \multicolumn{1}{c}{Label} & \multicolumn{2}{c}{2} & \multicolumn{2}{c}{5} & \multicolumn{2}{c}{10} & \multicolumn{2}{c}{2} & \multicolumn{2}{c}{5} & \multicolumn{2}{c}{10}\\
            \cmidrule(lr){2-3} \cmidrule(l){4-5} \cmidrule(l){6-7} \cmidrule(l){8-9} \cmidrule(l){10-11} \cmidrule(l){12-13}
             & Raw & OB & Raw & OB & Raw & OB & Raw & OB & Raw & OB & Raw & OB\\
            \midrule
            `Idle' & 46.41 & 49.52 & 42.97 & 41.03 & 44.0 & 37.79 & 41.59 & 53.24 & 48.34 & 49.79 & 40.48 & \textbf{60.48}\\
            `Leaving the house' & 81.62 & 81.33 & 80.72 & 71.73 & 82.64 & 56.73 & 81.65 & 83.55 & 85.32 & 80.09 & 78.74 & 65.5\\
            `Use toilet' & 2.6 & 1.3 & 0.0 & 0.0 & 0.0 & 0.0 & \textbf{6.49} & 2.6 & 0.0 & 2.6 & 0.0 & 0.0\\
            `Take shower' & 29.73 & 35.14 & 41.44 & 22.52 & 27.93 & 27.93 & 47.75 & 43.24 & 50.45 & 27.93 & 26.13 & 27.03\\
            `Brush teeth' & 0.0 & 0.0 & 0.0 & 0.0 & 0.0 & 0.0 & 0.0 & 0.0 & 0.0 & 0.0 & 0.0 & 0.0\\
            `Go to bed' & 92.31 & 94.98 & 91.5 & 95.42 & 92.09 & 89.13 & 91.0 & 95.04 & 91.3 & 96.11 & 92.12 & 86.48\\
            `Get dressed' & 19.57 & 26.09 & 39.13 & 23.91 & 36.96 & 4.35 & 30.43 & 36.96 & 34.78 & 30.43 & 45.65 & 23.91\\
            `Prepare brunch' & 16.67 & 5.95 & 3.57 & 4.76 & 13.1 & 20.24 & 14.29 & 9.52 & 23.81 & 9.52 & \textbf{33.33} & 20.24\\
            `Prepare dinner' & 33.8 & 36.62 & 32.39 & 45.07 & 12.68 & 39.44 & 32.39 & \textbf{60.56} & 50.7 & 54.93 & 49.3 & 42.25\\
            `Get a drink' & 0.0 & 0.0 & 0.0 & 0.0 & 0.0 & 0.0 & 0.0 & 0.0 & 0.0 & 0.0 & 0.0 & 0.0\\
            `Wash dishes' & 0.0 & 0.0 & 0.0 & 0.0 & 0.0 & 0.0 & 0.0 & 0.0 & 0.0 & 0.0 & 0.0 & 0.0\\
            `Eat dinner' & 0.0 & 0.0 & 0.0 & 0.0 & 0.0 & 0.0 & 0.0 & 0.0 & 0.0 & \textbf{22.86} & 0.0 & 0.0\\
            `Eat brunch' & 0.0 & 0.0 & 0.0 & 25.34 & 10.96 & 6.85 & 0.0 & 0.0 & 0.0 & 24.66 & 0.0 & 14.38\\
            \midrule
            Mean per class accuracy & $35.92$	& $36.65$	& $36.50$	& $35.78$ & $36.29$ & $30.81$ & $37.92$ & $38.70$ & $39.71$ & $39.45$ & $36.88$	& $36.08$\\
            Standard deviation & $13.11$	& $12.97$	& $11.74$	& $15.0$ & $13.21$ & $13.35$ & $11.86$ & $11.20$ & $10.90$ & $14.08$ & $11.73$	& $14.38$\\
            \midrule
            Overall accuracy & $80.27$ & $81.13$ & $79.27$ & $75.15$ & $80.58$ & $64.12$ & $79.67$ & $82.90$ & $82.49$ & $81.04$ & $78.19$ & $70.20$\\
            Standard deviation & $24.79$ & $20.34$ & $22.35$ & $22.36$ & $22.02$ & $21.68$ & $25.46$ & $20.13$ & $22.50$ & $20.70$ & $21.99$ & $23.73$\\
            \bottomrule
        \end{tabularx}
    \end{adjustbox}
    \label{labelacc_houseB_RawvsOB_LSTMGRU}
\end{table}

\begin{table}
    \caption{Accuracy per label using RNN and LSTMCRF (Raw vs OB feature representations) - House B (Look-back window: 2, 5 and 10)}
    \begin{adjustbox}{width=\columnwidth,center}
        \begin{tabularx}{1.6\textwidth}{c *{13}{C}}
            \toprule
            \multicolumn{1}{c}{} & \multicolumn{6}{c}{RNN} & \multicolumn{6}{c}{LSTMCRF}\\
            \cmidrule(lr){2-7} \cmidrule(l){8-13}
            \multicolumn{1}{c}{Label} & \multicolumn{2}{c}{2} & \multicolumn{2}{c}{5} & \multicolumn{2}{c}{10} & \multicolumn{2}{c}{2} & \multicolumn{2}{c}{5} & \multicolumn{2}{c}{10}\\
            \cmidrule(lr){2-3} \cmidrule(l){4-5} \cmidrule(l){6-7} \cmidrule(l){8-9} \cmidrule(l){10-11} \cmidrule(l){12-13}
             & Raw & OB & Raw & OB & Raw & OB & Raw & OB & Raw & OB & Raw & OB\\
            \midrule
            `Idle' & 35.52 & 46.28 & 38.41 & 39.59 & 38.62 & 43.79 & 39.86 & 54.97 & 57.31 & 42.83 & 51.79 & 47.45\\
            `Leaving the house' & 85.78 & 84.03 & 85.7 & 75.33 & 78.15 & 65.07 & 85.59 & \textbf{87.59} & 82.76 & 72.35 & 80.39 & 53.95\\
            `Use toilet' & 5.19 & 2.6 & 0.0 & \textbf{6.49} & 0.0 & 3.9 & 1.3 & 0.0 & 0.0 & 0.0 & 0.0 & 0.0\\
            `Take shower' & \textbf{60.36} & 44.14 & 45.95 & 32.43 & 50.45 & 27.03 & 21.62 & 13.51 & 36.94 & 0.0 & 22.52 & 0.0\\
            `Brush teeth' & 0.0 & 0.0 & 2.78 & \textbf{8.33} & 2.78 & 0.0 & 0.0 & 0.0 & 0.0 & 0.0 & 0.0 & 0.0\\
            `Go to bed' & 91.4 & 94.98 & 92.03 & \textbf{97.87} & 91.14 & 86.1 & 91.6 & 95.06 & 91.67 & 97.49 & 91.05 & 89.96\\
            `Get dressed' & 41.3 & 36.96 & \textbf{56.52} & 50.0 & 43.48 & 39.13 & 2.17 & 4.35 & 0.0 & 0.0 & 4.35 & 0.0\\
            `Prepare brunch' & 9.52 & 19.05 & 16.67 & 19.05 & 14.29 & 26.19 & 15.48 & 1.19 & 5.95 & 0.0 & 10.71 & 3.57\\
            `Prepare dinner' & 36.62 & 45.07 & 43.66 & 47.89 & 32.39 & 53.52 & 14.08 & 29.58 & 25.35 & 9.86 & 0.0 & 0.0\\
            `Get a drink' & 0.0 & 0.0 & 0.0 & 0.0 & 0.0 & 0.0 & 0.0 & 0.0 & 0.0 & 0.0 & 0.0 & 0.0\\
            `Wash dishes' & 0.0 & 0.0 & 0.0 & 0.0 & 0.0 & 0.0 & 0.0 & 0.0 & 0.0 & 0.0 & 0.0 & 0.0\\
            `Eat dinner' & 0.0 & 0.0 & 0.0 & 0.0 & 8.57 & 0.0 & 0.0 & 0.0 & 0.0 & 11.43 & 0.0 & 0.0\\
            `Eat brunch' & 0.68 & 8.22 & 0.0 & 6.85 & 17.81 & \textbf{27.4} & 0.0 & 0.0 & 0.0 & 24.66 & 2.05 & 24.66\\
            \midrule
            Mean per class accuracy & $39.29$	& $39.49$ & $40.16$ & $\mathbf{40.28}$ & $39.02$ & $34.96$ & $33.61$ & $34.60$ & $34.77$ & $32.45$ & $32.86$ & $28.48$\\
            Standard deviation & $10.38$	& $10.82$ & $9.81$ & $13.37$ & $12.23$ & $10.12$ & $14.53$ & $13.60$ & $13.34$ & $15.23$ & $13.80$ & $13.50$\\
            \midrule
            Overall accuracy & $81.86$ & $82.70$ & $82.19$ & $77.98$ & $77.56$ & $68.84$ & $81.76$ & $\mathbf{85.0}$ & $81.44$ & $75.95$ & $79.34$ & $63.18$\\
            Standard deviation & $22.27$ & $20.10$ & $21.89$ & $21.93$ & $25.43$ & $21.24$ & $22.29$ & $16.38$ & $22.38$ & $19.94$ & $22.21$ & $18.34$\\
            \bottomrule
        \end{tabularx}
    \end{adjustbox}
    \label{labelacc_houseB_RawvsOB_RNNLSTMCRF}
\end{table}

For house B, LSTMCRF is the method which provides highest accuracy considering a look-back window of $2$. As for the mean per class accuracy, RNN with a look-back window of $5$ is the method that performs the best, but we observe once again that there is not a significant difference between the mean per class accuracies for a look-back window of $2$ and $5$.

\begin{table}
    \caption{Accuracy per label using LSTM and GRU (Raw vs OB feature representations) - House C (Look-back window: 2, 5 and 10)}
    \begin{adjustbox}{width=\columnwidth,center}
        \begin{tabularx}{1.6\textwidth}{c *{13}{C}}
            \toprule
            \multicolumn{1}{c}{} & \multicolumn{6}{c}{LSTM} & \multicolumn{6}{c}{GRU}\\
            \cmidrule(lr){2-7} \cmidrule(l){8-13}
            \multicolumn{1}{c}{Label} & \multicolumn{2}{c}{2} & \multicolumn{2}{c}{5} & \multicolumn{2}{c}{10} & \multicolumn{2}{c}{2} & \multicolumn{2}{c}{5} & \multicolumn{2}{c}{10}\\
            \cmidrule(lr){2-3} \cmidrule(l){4-5} \cmidrule(l){6-7} \cmidrule(l){8-9} \cmidrule(l){10-11} \cmidrule(l){12-13}
             & Raw & OB & Raw & OB & Raw & OB & Raw & OB & Raw & OB & Raw & OB\\
            \midrule
            `Idle' & 38.5 & 51.93 & 49.21 & \textbf{66.35} & 48.32 & 64.0 & 34.05 & 53.22 & 45.73 & 52.38 & 35.04 & 56.91\\
            `Leave house' & 32.72 & 42.45 & 24.6 & 32.64 & 25.71 & 14.25 & 24.03 & 45.13 & 26.85 & 41.1 & 28.05 & 27.6\\
            `Eating' & 2.24 & 3.99 & 8.73 & 20.95 & 22.69 & 12.22 & 2.49 & 3.99 & 11.22 & 17.96 & 18.45 & \textbf{28.18}\\
            `Use toilet downstairs' & 0.0 & 0.0 & 0.0 & 0.0 & 0.0 & 0.0 & 0.0 & 0.0 & 0.0 & 0.0 & 0.0 & 0.0\\
            `Take shower' & 0.0 & 3.68 & 4.74 & 14.74 & 2.11 & 0.0 & 0.53 & 6.84 & 2.63 & 7.37 & 2.63 & 8.95\\
            `Brush teeth' & 1.98 & 0.0 & 0.0 & 0.0 & 0.0 & 0.0 & 2.97 & 0.0 & 1.98 & 0.0 & 0.0 & 0.0\\
            `Use toilet upstairs' & 8.75 & 1.25 & 1.25 & 0.0 & 0.0 & 0.0 & 11.25 & 5.0 & 2.5 & 0.0 & 1.25 & 0.0\\
            `Shave' & 0.0 & 0.0 & 2.9 & 0.0 & 0.0 & 0.0  & 0.0 & 1.45 & 1.45 & 0.0 & 0.0 & 0.0\\
            `Go to bed' & 68.39 & 96.26 & 74.87 & 96.39 & 78.09 & 88.53 & 70.11 & 93.43 & 74.68 & 97.19 & 75.36 & 91.92\\
            `Get dressed' & 7.14 & 7.14 & 7.14 & 19.64 & 12.5 & 10.71 & 8.04 & 15.18 & 9.82 & 20.54 & 4.46 & 25.0\\
            `Take medication' & 0.0 & 0.0 & 0.0 & 0.0 & 0.0 & 0.0 & 0.0 & 0.0 & \textbf{6.67} & 0.0 & 0.0 & 0.0\\
            `Prepare breakfast' & 0.0 & 0.0 & 0.0 & 0.0 & 0.0 & 0.0 & 0.0 & 0.0 & 0.0 & 2.82 & 0.0 & 0.0\\
            `Prepare lunch' & 0.0 & 0.0 & 0.0 & 0.0 & 0.0 & 0.0 & 0.0 & 0.0 & 0.0 & 0.0 & 0.0 & 0.0\\
            `Prepare dinner' & 7.24 & 8.28 & 9.31 & 5.86 & 5.86 & 4.48 & 7.59 & 9.31 & 10.0 & 15.52 & 8.28 & 8.97\\
            `Get snack' & 0.0 & 0.0 & 0.0 & 0.0 & 0.0 & 0.0 & 0.0 & 0.0 & 0.0 & 0.0 & 0.0 & 0.0\\
            `Get drink' & 0.0 & 0.0 & 0.0 & 0.0 & 0.0 & 0.0  & 0.0 & 0.0 & 0.0 & 0.0 & 0.0 & 0.0\\
            \midrule
            Mean per class accuracy & $15.70$	& $23.09$	& $17.69$ & $26.52$ & $18.88$ & $21.37$ & $15.06$ & $24.68$ & $18.22$ & $26.93$ & $16.43$ & $26.56$\\
            Standard deviation & $8.09$	& $19.98$	& $8.15$ & $19.15$ & $10.22$ & $19.70$ & $8.26$ & $19.95$ & $11.12$ & $18.76$ & $7.02$ & $19.17$\\
            \midrule
            Overall accuracy & $42.33$ & $59.31$ & $42.58$ & $58.04$ & $44.42$ & $47.06$ & $38.21$ & $60.01$ & $43.33$ & $59.43$ & $41.22$ & $52.96$\\
            Standard deviation & $19.44$ & $19.92$ & $19.62$ & $22.98$ & $21.19$ & $19.45$ & $19.03$ & $21.83$ & $ 23.13$ & $20.74$ & $21.38$ & $20.72$\\
            \bottomrule
        \end{tabularx}
    \end{adjustbox}
    \label{labelacc_houseC_RawvsOB_LSTMGRU}
\end{table}

\begin{table}
    \caption{Accuracy per label using RNN and LSTMCRF (Raw vs OB feature representations) - House C (Look-back window: 2, 5 and 10)}
    \begin{adjustbox}{width=\columnwidth,center}
        \begin{tabularx}{1.6\textwidth}{c *{13}{C}}
            \toprule
            \multicolumn{1}{c}{} & \multicolumn{6}{c}{RNN} & \multicolumn{6}{c}{LSTMCRF}\\
            \cmidrule(lr){2-7} \cmidrule(l){8-13}
            \multicolumn{1}{c}{Label} & \multicolumn{2}{c}{2} & \multicolumn{2}{c}{5} & \multicolumn{2}{c}{10} & \multicolumn{2}{c}{2} & \multicolumn{2}{c}{5} & \multicolumn{2}{c}{10}\\
            \cmidrule(lr){2-3} \cmidrule(l){4-5} \cmidrule(l){6-7} \cmidrule(l){8-9} \cmidrule(l){10-11} \cmidrule(l){12-13}
             & Raw & OB & Raw & OB & Raw & OB & Raw & OB & Raw & OB & Raw & OB\\
            \midrule
            `Idle' & 33.13 & 61.62 & 40.49 & 64.72 & 41.12 & 57.25 & 33.85 & 49.39 & 48.05 & 61.78 & 49.31 & 60.37\\
            `Leave house' & 24.64 & 47.19 & 25.07 & 19.94 & 21.19 & 22.85 & 41.53 & \textbf{50.81} & 28.54 & 31.26 & 10.98 & 12.59\\
            `Eating' & 3.49 & 4.24 & 9.73 & 23.94 & 19.45 & 18.7 & 3.49 & 0.25 & 9.73 & 17.21 & 21.95 & 12.72\\
            `Use toilet downstairs' & 0.0 & 0.0 & 0.0 & 0.0 & 0.0 & 0.0 & 0.0 & 0.0 & 0.0 & 0.0 & 0.0 & 0.0\\
            `Take shower' & 2.63 & 7.89 & 7.37 & \textbf{17.37} & 7.89 & 7.89 & 0.53 & 5.79 & 2.63 & 8.42 & 4.74 & 2.11\\
            `Brush teeth' & 3.96 & 0.0 & 0.99 & \textbf{4.95} & 3.96 & 1.98 & 2.97 & 0.0 & 0.0 & 0.0 & 0.0 & 0.0\\
            `Use toilet upstairs' & 7.5 & 2.5 & 6.25 & 0.0 & 1.25 & 1.25 & \textbf{13.75} & 0.0 & 2.5 & 0.0 & 0.0 & 0.0\\
            `Shave' & 0.0 & \textbf{10.14} & 0.0 & 1.45 & 0.0 & 0.0  & 1.45 & 0.0 & 0.0 & 0.0 & 0.0 & 0.0 \\
            `Go to bed' & 74.02 & \textbf{98.35} & 79.63 & 92.4 & 79.08 & 86.44 & 68.92 & 96.25 & 74.55 & 91.43 & 80.82 & 87.55\\
            `Get dressed' & 10.71 & 18.75 & 11.61 & \textbf{33.04} & 14.29 & 19.64 & 10.71 & 4.46 & 8.93 & 20.54 & 13.39 & 8.93\\
            `Take medication' & 0.0 & 0.0 & 0.0 & 0.0 & 0.0 & 0.0 & 0.0 & 0.0 & 0.0 & 0.0 & 0.0 & 0.0\\
            `Prepare breakfast' & \textbf{5.63} & 1.41 & 0.0 & 2.82 & 2.82 & 0.0 & 0.0 & 0.0 & 0.0 & 0.0 & 0.0 & 0.0\\
            `Prepare lunch' & 0.0 & 0.0 & \textbf{1.67} & 0.0 & 0.0 & 0.0 & 0.0 & 0.0 & 0.0 & 0.0 & 0.0 & 0.0\\
            `Prepare dinner' & 5.52 & 3.45 & 3.1 & 14.83 & 2.41 & 8.97 & 7.93 & \textbf{16.55} & 7.93 & 7.24 & 3.79 & 0.0\\
            `Get snack' & 0.0 & 0.0 & 0.0 & 0.0 & 0.0 & 0.0  & 0.0 & 0.0 & 0.0 & 0.0 & 0.0 & 0.0\\
            `Get drink' & 0.0 & 0.0 & 0.0 & 0.0 & 0.0 & 0.0  & 0.0 & 0.0 & 0.0 & 0.0 & 0.0 & 0.0\\
            \midrule
            Mean per class accuracy & $19.35$ & $25.66$ & $21.68$	& $\mathbf{28.07}$ & $21.54$ & $19.86$ & $15.19$ & $23.47$ & $17.01$ & $25.41$ & $19.67$ & $20.25$\\
            Standard deviation & $20.14$ & $19.31$ & $20.13$	& $18.80$ & $20.36$ & $6.37$ & $7.86$ & $19.31$ & $8.30$ & $19.68$ & $19.92$ & $19.74$\\
            \midrule
            Overall accuracy & $41.84$ & $\mathbf{63.80}$ & $44.97$ & $51.29$ & $43.25$ & $45.67$ & $45.67$ & $62.53$ & $44.09$ & $55.12$ & $40.96$ & $45.36$\\
            Standard deviation & $24.28$ & $19.29$ & $23.65$ & $22.47$ & $25.23$ & $18.05$ & $23.23$ & $20.32$ & $22.56$ & $23.0$ & $21.90$ & $18.58$\\
            \bottomrule
        \end{tabularx}
    \end{adjustbox}
    \label{labelacc_houseC_RawvsOB_RNNLSTMCRF}
\end{table}

Lastly, for house C, the RNN method achieved the highest values for the evaluation metrics considered, where a look-back window of $5$ and $2$ gave the best results for the mean per-class accuracy and the accuracy, respectively.

In general, we observe that for lower look-back window values, our proposed feature representation achieves significantly better results than the raw representation. Moreover, from the results obtained for the LSTM, GRU, RNN and LSTMCRF models, we conclude that the best accuracy for all houses was obtained by considering a look-back window of $2$ and an OB feature representation of the data. In addition, neural network models seem not to benefit much from concatenating multiple data points for training as those techniques learn temporal dependencies differently.

We also note that, if considering the raw feature representation, a longer look-back window is required so that LSTM models are able to obtain reasonable results. In particular, it becomes hard to accurately predict labels due to the long-term dependencies inherent to the raw feature representation. Therefore, based on the results obtained, this implies that there is an advantage in using the proposed feature representation. The OB feature representation is shown to be beneficial not only in obtaining a higher accuracy but also in decreasing the training time given that a better performance is achieved when considering a low look-back window value.

\subsubsection{Observation-based representation with probabilistic-based methods and a MLP network model}

In this experiment, we use probabilistic models and a feed forward neural network model and considered an OB feature representation. Unlike recurrent neural networks, models such as NB, HMM, HSMM, CRF and MLP are limited to a single ``time step" (i.e. a look-back window of $1$). However, it is possible to provide look-back information to these models. We accomplish this by feeding in a sequence which contains concatenated data points. Specifically, we add the most recent data points as further features of the current single data point. We consider $2$, $5$ and $10$ as the possible values for the number of recent data points to be concatenated with the current one.

Also, we do not consider the raw representation for these models as it would result in low information signals, where repeated information would be given as input to the models in the form of equal concatenated data points.

For both overall accuracy as well as per-class accuracies, CRFs were able to outperform all the experiments done thus far by using an OB feature representation (Tables~\ref{labelacc_houseA}, \ref{labelacc_houseB} and \ref{labelacc_houseC}). The best accuracy values were obtained by concatenating $5$ data points for house A (97.14$\pm$5.89) and $10$ data points for houses B (87.55$\pm$16.77) and C (90.43$\pm$14.85). Nevertheless, the experiments also show that a higher value for the number of concatenated data points significantly contributes towards a higher mean per class accuracy.

\begin{table}
    \caption{Accuracy per label (OB feature representation) - House A (Data points concatenated: 2, 5 and 10)}
    \begin{adjustbox}{width=\columnwidth,center}
        \begin{tabularx}{1.82\textwidth}{c *{16}{C}}
            \toprule
            \multicolumn{1}{c}{Label} & \multicolumn{3}{c}{NB} & \multicolumn{3}{c}{HMM} & \multicolumn{3}{c}{HSMM} & \multicolumn{3}{c}{CRF} & \multicolumn{3}{c}{MLP}\\
            \cmidrule(lr){2-4} \cmidrule(l){5-7} \cmidrule(l){8-10} \cmidrule(l){11-13} \cmidrule(l){14-16}
             & 2 & 5 & 10 & 2 & 5 & 10 & 2 & 5 & 10 & 2 & 5 & 10 & 2 & 5 & 10\\
            \midrule
            `Idle' & 84.38 & 57.81 & 43.81 & 51.41 & 33.59 & 22.2 & 53.58 & 36.8 & 22.87 & 84.34 & \textbf{92.58} & 91.99 & 86.5 & 85.1 &71.71\\
            `Leave house' & 99.48 & 97.84 & 82.56 & 94.79 & 78.47 & 57.22 & 94.67 & 85.92 & 72.61 & \textbf{99.9} & 99.91 & 96.74 & \textbf{99.9} & 99.81 & 94.33\\
            `Use toilet' & 63.84 & 57.81 & 51.23 & 69.32 & 51.51 & 49.86 & 72.33 & 58.08 & 51.51 & 56.44 & 69.04 & \textbf{82.74} & 69.04 & 64.93 & 61.37\\
            `Take shower' & 0.0 & 7.57 & 33.47 & 54.98 & 43.03 & 49.4 & 61.35 & 35.46 & 52.99 & 17.53 & 60.16 & \textbf{77.69} & 0.0 & 4.38 & 19.52\\
            `Brush teeth' & 0.0 & 9.38 & 15.62 & 28.12 & \textbf{46.88} & 37.5 & 31.25 & 37.5 & 28.12 & 0.0 & 12.5 & \textbf{46.88} & 0.0 & 0.0 & 0.0\\
            `Go to bed' & 97.75 & 89.74 & 82.79 & 88.57 & 86.74 & 74.67 & 89.8 & 89.15 & 76.06 & 98.25 & 96.57 & 97.7 & \textbf{98.97} & 97.79 & 92.34\\
            `Prepare breakfast' & 44.83 & 44.83 & 54.02 & 50.57 & 51.72 & 60.92 & 49.43 & 54.02 & 59.77 & 66.67 & \textbf{86.21} & \textbf{86.21} & 52.87 & 52.87 & 58.62\\
            `Prepare dinner' & 13.59 & 19.16 & 17.42 & 57.14 & 49.48 & 39.72 & 56.79 & 39.02 & 39.72 & 75.96 & 88.5 & \textbf{96.17} & 11.15 & 18.12 & 21.6\\
            `Get snack' & 23.81 & 40.48 & 40.48 & 64.29 & 42.86 & 40.48 & 64.29 & 42.86 & 42.86 & 47.62 & 83.33 & \textbf{97.62} & 4.76 & 4.76 & 7.14\\
            `Get drink' & 36.73 & 32.65 & 32.65 & 34.69 & 24.49 & 34.69 & 38.78 & 38.78 & 34.69 & \textbf{81.63} & \textbf{81.63} & \textbf{81.63} & 48.98 & 53.06 & 51.02\\
            \midrule
            Mean per class accuracy & $61.67$	& $57.52$	& $55.86$	& $67.79$ & $55.46$ & $49.71$ & $70.18$ & $60.18$ & $54.68$ & $69.84$ & $80.26$ & $\mathbf{86.7}$ & $63.25$ & $64.25$ & $62.64$\\
            Standard deviation & $15.29$ & $17.55$	& $20.53$	& $17.78$ & $21.63$ & $24.84$ & $17.82$ & $17.73$ & $23.14$ & $13.69$ & $12.65$ & $12.57$ & $14.42$ & $14.10$ & $17.53$\\
            \midrule
            Overall accuracy & $95.0$ & $88.65$ & $76.73$ & $86.71$ & $74.86$ & $58.46$ & $87.37$ & $79.91$ & $67.26$ & $96.04$ & $\mathbf{97.14}$ & $96.12$ & $95.89$ & $95.37$ & $89.33$\\
            Standard deviation & $3.58$ & $9.01$ & $24.12$ & $13.64$ & $28.48$ & $30.51$ & $13.27$ & $22.41$ & $25.8$ & $2.92$ & $5.89$ & $8.29$ & $2.62$ & $3.54$ & $11.74$\\
            \bottomrule
        \end{tabularx}
    \end{adjustbox}
    \label{labelacc_houseA}
\end{table}
\begin{table}
    \caption{Accuracy per label (OB feature representation) - House B (Data points concatenated: 2, 5 and 10)}
    \begin{adjustbox}{width=\columnwidth,center}
        \begin{tabularx}{1.9\textwidth}{c *{16}{C}}
            \toprule
            \multicolumn{1}{c}{Label} & \multicolumn{3}{c}{NB} & \multicolumn{3}{c}{HMM} & \multicolumn{3}{c}{HSMM} & \multicolumn{3}{c}{CRF} & \multicolumn{3}{c}{MLP}\\
            \cmidrule(lr){2-4} \cmidrule(l){5-7} \cmidrule(l){8-10} \cmidrule(l){11-13} \cmidrule(l){14-16}
             & 2 & 5 & 10 & 2 & 5 & 10 & 2 & 5 & 10 & 2 & 5 & 10 & 2 & 5 & 10\\
            \midrule
            `Idle' & 42.55 & 35.93 & 44.14 & 21.52 & 25.03 & 32.41 & 23.52 & 25.52 & 34.21 & 55.66 & 51.1 & \textbf{60.9} & 45.52 & 40.28 & 40.34\\
            `Leaving the house' & 85.15 & 65.57 & 56.01 & 61.27 & 59.71 & 54.75 & 61.27 & 59.71 & 54.75 & \textbf{92.7} & 83.7 & 90.45 & 88.03 & 78.34 & 75.11\\
            `Use toilet' & 15.58 & 15.58 & 14.29 & 31.17 & 16.88 & 20.78 & 31.17 & 20.78 & 20.78 & 16.88 & 37.66 & \textbf{44.16} & 0.0 & 6.49 & 6.49\\
            `Take shower' & 35.14 & 15.32 & 21.62 & 54.95 & 45.95 & 18.02 & 55.86 & 45.95 & 18.02 & 64.86 & 63.06 & \textbf{72.07} & 32.43 & 34.23 & 28.83\\
            `Brush teeth' & 0.0 & 8.33 & 16.67 & 13.89 & 13.89 & 16.67 & 8.33 & 13.89 & 16.67 & 36.11 & 25.0 & \textbf{80.56} & 0.0 & 0.0 & 0.0\\
            `Go to bed' & 81.77 & 77.53 & 73.06 & 79.29 & 73.81 & 72.55 & 80.09 & 73.76 & 72.55 & 87.27 & 86.21 & 89.96 & \textbf{92.83} & 90.58 & 88.1\\
            `Get dressed' & 50.0 & 56.52 & 43.48 & 58.7 & 67.39 & 45.65 & 56.52 & 65.22 & 45.65 & 71.74 & 67.39 & \textbf{80.43} & 19.57 & 41.3 & 36.96\\
            `Prepare brunch' & 21.43 & 23.81 & 21.43 & 27.38 & 23.81 & 21.43 & 26.19 & 23.81 & 21.43 & \textbf{78.57} & 70.24 & 75.0 & 19.05 & 23.81 & 25.0\\
            `Prepare dinner' & 30.99 & 25.35 & 26.76 & 33.8 & 28.17 & 32.39 & 33.8 & 28.17 & 32.39 & \textbf{97.18} & 95.77 & 94.37 & 21.13 & 14.08 & 12.68\\
            `Get a drink' & 0.0 & 0.0 & 0.0 & 0.0 & 0.0 & 0.0 & 14.29 & 0.0 & 0.0 & 14.29 & 14.29 & \textbf{28.57} & 0.0 & 0.0 & 0.0\\
            `Wash dishes' & 0.0 & 0.0 & 0.0 & 0.0 & 0.0 & 0.0 & 0.0 & 0.0 & 0.0 & \textbf{47.62} & 4.76 & 9.52 & 0.0 & 0.0 & 0.0\\
            `Eat dinner' & 17.14 & 0.0 & 0.0 & 0.0 & 0.0 & 0.0 & 0.0 & 0.0 & 0.0 & 80.0 & 80.0 & \textbf{100.0} & 0.0 & 0.0 & 0.0\\
            `Eat brunch' & 10.27 & 15.07 & 16.44 & 20.55 & 16.44 & 28.08 & 20.55 & 16.44 & 28.08 & 8.22 & \textbf{61.64} & 60.96 & 2.05 & 6.85 & 6.85\\
            \midrule
            Mean per class accuracy & $35.85$	& $32.15$	& $30.54$	& $36.26$ & $36.15$ & $31.77$ & $36.42$ & $36.38$ & $31.90$ & $57.24$ & $61.97$  & $\mathbf{71.08}$ & $46.06$ & $46.44$ & $45.34$\\
            Standard deviation & $8.42$	& $10.75$	& $11.28$	& $11.13$ & $11.79$ & $13.38$ & $12.61$ & $11.73$ & $13.37$ & $18.52$ & $23.61$  & $23.40$ & $15.42$ & $16.13$ & $17.10$\\
            \midrule
            Overall accuracy & $79.17$ & $65.85$ & $59.49$ & $63.19$ & $60.74$ & $57.87$ & $63.57$ & $60.77$ & $58.00$ & $87.10$ & $81.51$ & $\mathbf{87.55}$ & $85.62$ & $81.91$ & $80.41$\\
            Standard deviation & $19.03$ & $27.03$ & $27.8$ & $25.02$ & $28.96$ & $27.86$ & $24.66$ & $28.90$ & $27.77$ & $22.02$ & $23.85$ & $16.77$ & $16.44$ & $20.0$ & $20.90$\\
            \bottomrule
        \end{tabularx}
    \end{adjustbox}
    \label{labelacc_houseB}
\end{table}
\begin{table}[H]
    \caption{Accuracy per label (OB feature representation) - House C (Data points concatenated: 2, 5 and 10)}
    \begin{adjustbox}{width=\columnwidth,center}
        \begin{tabularx}{1.82\textwidth}{c *{16}{C}}
            \toprule
            \multicolumn{1}{c}{Label} & \multicolumn{3}{c}{NB} & \multicolumn{3}{c}{HMM} & \multicolumn{3}{c}{HSMM} & \multicolumn{3}{c}{CRF} & \multicolumn{3}{c}{MLP}\\
            \cmidrule(lr){2-4} \cmidrule(l){5-7} \cmidrule(l){8-10} \cmidrule(l){11-13} \cmidrule(l){14-16}
             & 2 & 5 & 10 & 2 & 5 & 10 & 2 & 5 & 10 & 2 & 5 & 10 & 2 & 5 & 10\\
            \midrule
            `Idle' & 43.66 & 38.52 & 33.77 & 27.4 & 14.99 & 20.16 & 33.33 & 17.12 & 20.63 & 67.71 & 68.31 & \textbf{81.54} & 31.79 & 50.87 & 47.4\\
            `Leave house' & 43.31 & 29.34 & 24.79 & 23.41 & 19.25 & 13.81 & 23.34 & 19.27 & 13.69 & 91.32 & 90.07 & \textbf{97.34} & 58.87 & 36.42 & 35.0\\
            `Eating' & 5.24 & 12.22 & 23.44 & 4.24 & 14.21 & 21.45 & 4.24 & 15.71 & 20.95 & 15.96 & 40.4 & \textbf{46.38} & 4.99 & 14.71 & 20.95\\
            `Use toilet downstairs' & 0.0 & 0.0 & 0.0 & 2.53 & 5.7 & 2.53 & 2.53 & 4.43 & 2.53 & 0.0 & 1.27 & \textbf{10.13} & 0.0 & 0.0 & 0.0\\
            `Take shower' & 3.68 & 5.79 & 2.63 & 13.16 & 5.79 & 2.63 & 12.63 & 5.79 & 2.63 & \textbf{62.11} & 61.05 & 50.0 & 5.26 & 13.68 & 10.0\\
            `Brush teeth' & 14.85 & 17.82 & 23.76 & 25.74 & 29.7 & 25.74 & 24.75 & \textbf{32.67} & 25.74 & 12.87 & 23.76 & 25.74 & 0.99 & 3.96 & 7.92\\
            `Use toilet upstairs' & 17.5 & 17.5 & 23.75 & 25.0 & 21.25 & 28.75 & 25.0 & 23.75 & \textbf{30.0} & 15.0 & 28.75 & \textbf{30.0} & 8.75 & 7.5 & 8.75\\
            `Shave' & 10.14 & 10.14 & 13.04 & 13.04 & 13.04 & 13.04 & 15.94 & 13.04 & 13.04 & 30.43 & 62.32 & \textbf{85.51} & 0.0 & 0.0 & 0.0\\
            `Go to bed' & 79.22 & 71.83 & 48.23 & 59.85 & 44.73 & 22.89 & 59.38 & 44.71 & 26.71 & 93.44 & 91.1 & 92.78 & 93.35 & 92.94 & \textbf{96.84}\\
            `Get dressed' & 22.32 & 26.79 & 19.64 & 46.43 & 33.04 & 18.75 & 43.75 & 32.14 & 22.32 & 12.5 & \textbf{53.57} & 52.68 & 10.71 & 23.21 & 23.21\\
            `Take medication' & 13.33 & 0.0 & 13.33 & 0.0 & 0.0 & 13.33 & 0.0 & 0.0 & 13.33 & 13.33 & 13.33 & \textbf{46.67} & 0.0 & 0.0 & 0.0\\
            `Prepare breakfast' & 0.0 & 5.63 & 4.23 & 8.45 & 8.45 & 11.27 & 8.45 & 8.45 & 11.27 & 40.85 & 50.7 & \textbf{78.87} & 0.0 & 2.82 & 1.41\\
            `Prepare lunch' & 0.0 & 5.0 & 0.0 & 8.33 & 23.33 & 0.0 & 15.0 & 20.0 & 0.0 & 3.33 & \textbf{25.0} & 18.33 & 0.0 & 0.0 & 0.0\\
            `Prepare dinner' & 16.55 & 23.45 & 2.07 & 14.48 & 18.62 & 0.0 & 12.41 & 18.62 & 0.0 & 60.69 & 72.76 & \textbf{80.34} & 5.17 & 24.14 & 3.79\\
            `Get snack' & 0.0 & 0.0 & 12.5 & 8.33 & \textbf{16.67} & 12.5 & 8.33 & \textbf{16.67} & 12.5 & 0.0 & 12.5 & 8.33 & 0.0 & 0.0 & 0.0\\
            `Get drink' & 0.0 & 0.0 & 0.0 & 0.0 & 0.0 & 0.0 & 0.0 & 0.0 & 0.0 & 0.0 & \textbf{45.16} & 25.81 & 0.0 & 0.0 & 0.0\\
            \midrule
            Mean per class accuracy & $20.09$	& $21.0$	& $17.23$	& $21.61$ & $19.05$ & $15.04$ & $22.14$ & $19.48$ & $15.86$ & $38.74$ & $49.10$	& $\mathbf{55.62}$ & $32.20$ & $34.83$ & $34.62$\\
            Standard deviation & $8.27$	& $8.43$	& $9.66$	& $7.88$ & $8.05$ & $10.48$ & $8.72$ & $ 7.99$ & $10.44$ & $17.13$ & $17.34$	& $16.57$ & $23.63$ & $22.11$ & $20.82$\\
            \midrule
            Overall accuracy & $49.61$ & $40.93$ & $31.19$ & $32.76$ & $24.59$ & $16.82$ & $33.67$ & $25.0$ & $17.93$ & $84.26$ & $84.12$ & $\mathbf{90.43}$ & $69.46$ & $64.90$ & $64.37$\\
            Standard deviation & $21.53$ & $18.83$ & $19.34$ & $20.74$ & $18.36$ & $19.13$ & $20.46$ & $18.05$ & $19.02$ & $14.05$ & $20.48$ & $14.85$ & $23.81$ & $25.74$ & $26.11$\\
            \bottomrule
        \end{tabularx}
    \end{adjustbox}
    \label{labelacc_houseC}
\end{table}

\subsection{Adding the Time of Day as a further feature}

From the results presented in the last section, it is possible to conclude that CRF is the algorithm which overwhelmingly is able to perform the best using an OB feature representation. In this section, we show the results obtained by adding the time of day ($hour$)  as a further feature to the dataset. In total, we considered fifteen different feature combinations in our experiments.

\begin{table}
    \caption{Accuracy using CRF (OB feature representation) - Houses A, B and C}
    \begin{adjustbox}{width=\columnwidth,center}
        \begin{tabularx}{1.5\textwidth}{c *{10}{C}}
            \toprule
            \multicolumn{1}{c}{Feature} & \multicolumn{3}{c}{House A} & \multicolumn{3}{c}{House B} & \multicolumn{3}{c}{House C}\\
            \cmidrule(lr){2-4} \cmidrule(l){5-7} \cmidrule(l){8-10}
            \multicolumn{1}{c}{Combination} & 2 & 5 & 10 & 2 & 5 & 10 & 2 & 5 & 10\\
            \midrule
             &  &  &  &  &  &  &  &  & \\
            NoToD\&NoDeltaT & $96.04$ & $97.14$ & $96.12$ & $87.10$ & $81.51$ & $87.55$ & $84.26$ & $84.12$ & $90.43$\\
            Standard deviation & $2.92$ & $5.89$ & $8.29$ & $22.02$ & $23.85$ & $16.77$ & $14.05$ & $20.48$ & $14.85$\\
             &  &  &  &  &  &  &  &  & \\
            NoToD\&OneHotDeltaT7 & $97.62$ & $98.70$ & $98.05$ & $87.21$ & $90.25$ & $88.47$ & $84.39$ & $81.98$ & $91.07$\\
            Standard deviation & $2.01$ & $1.66$ & $3.56$ & $16.7$ & $11.74$ & $15.38$ & $15.49$ & $21.87$ & $13.82$\\
             &  &  &  &  &  &  &  &  & \\
            NoToD\&OneHotDeltaT48 & $98.29$ & $98.10$ & $98.77$ & $88.28$ & $90.89$ & $89.77$ & $85.97$ & $90.38$ & $92.43$\\
            Standard deviation & $1.17$ & $4.69$ & $3.26$ & $13.45$ & $11.44$ & $15.61$ & $13.42$ & $17.03$ & $16.5$\\
             &  &  &  &  &  &  &  &  & \\
            NoToD\&UnaryDeltaT7 & $97.55$ & $98.49$ & $98.55$ & $88.49$ & $87.27$ & $91.96$ & $83.28$ & $86.63$ & $89.12$\\
            Standard deviation & $1.88$ & $1.97$ & $2.98$ & $13.61$ & $18.71$ & $14.63$ & $15.89$ & $17.22$ & $16.89$\\
             &  &  &  &  &  &  &  &  & \\
            NoToD\&UnaryDeltaT48 & $98.32$ & $98.41$ & $98.79$ & $89.59$ & $92.13$ & $88.85$ & $88.88$ & $86.78$ & $91.21$\\
            Standard deviation & $1.15$ & $3.54$ & $3.37$ & $19.98$ & $11.81$ & $14.14$ & $11.53$ & $19.26$ & $16.98$\\
             &  &  &  &  &  &  &  &  & \\
            \midrule
             &  &  &  &  &  &  &  &  & \\
            OneHotToD\&NoDeltaT & $97.78$ & $98.24$ & $96.85$ & $86.53$ & $91.80$ & $87.66$ & $87.01$ & $88.84$ & $90.82$\\
            Standard deviation & $1.88$ & $3.22$ & $7.91$ & $20.65$ & $12.39$ & $17.12$ & $11.89$ & $17.43$ & $15.02$\\
             &  &  &  &  &  &  &  &  & \\
            OneHotToD\&OneHotDeltaT7 & $98.29$ & $98.34$ & $97.85$ & $90.95$ & $92.65$ & $90.53$ & $88.27$ & $89.22$ & $86.88$\\
            Standard deviation & $1.36$ & $4.03$ & $5.77$ & $12.29$ & $8.85$ & $12.99$ & $13.80$ & $17.25$ & $19.42$\\
             &  &  &  &  &  &  &  &  & \\
            OneHotToD\&OneHotDeltaT48 & $98.16$ & $98.50$ & $95.94$ & $88.18$ & $92.72$ & $91.23$ & $84.52$ & $92.74$ & $92.88$\\
            Standard deviation & $3.30$ & $3.71$ & $11.82$ & $17.04$ & $12.68$ & $17.02$ & $20.44$ & $13.24$ & $14.39$\\
             &  &  &  &  &  &  &  &  & \\
            OneHotToD\&UnaryDeltaT7 & $98.37$ & $98.31$ & $98.17$ & $91.12$ & $95.47$ & $94.11$ & $83.64$ & $88.10$ & $88.81$\\
            Standard deviation & $1.38$ & $2.57$ & $4.64$ & $12.56$ & $6.73$ & $8.13$ & $17.20$ & $16.74$ & $15.89$\\
             &  &  &  &  &  &  &  &  & \\
            OneHotToD\&UnaryDeltaT48 & $98.71$ & $98.77$ & $98.79$ & $87.97$ & $93.72$ & $93.05$ & $89.36$ & $91.72$ & $\mathbf{94.10}$\\ 
            Standard deviation & $1.39$ & $3.29$ & $2.68$ & $16.53$ & $9.80$ & $13.39$ & $15.27$ & $15.36$ & $15.27$\\ 
             &  &  &  &  &  &  &  &  & \\
            \midrule
             &  &  &  &  &  &  &  &  & \\
            UnaryToD\&NoDeltaT & $97.90$ & $97.00$ & $97.40$ & $87.15$ & $90.35$ & $95.48$ & $87.54$ & $89.78$ & $92.77$\\
            Standard deviation & $1.70$ & $5.20$ & $5.93$ & $19.28$ & $16.93$ & $8.27$ & $14.53$ & $16.42$ & $11.88$\\
             &  &  &  &  &  &  &  &  & \\
            UnaryToD\&OneHotDeltaT7 & $97.92$ & $98.49$ & $98.38$ & $93.23$ & $95.63$ & $92.89$ & $87.48$ & $86.06$ & $86.43$\\
            Standard deviation & $2.81$ & $3.09$ & $3.90$ & $7.91$ & $8.23$ & $14.50$ & $16.62$ & $16.53$ & $18.40$\\
             &  &  &  &  &  &  &  &  & \\
            UnaryToD\&OneHotDeltaT48 & $98.69$ & $97.66$ & $96.00$ & $93.95$ & $94.63$ & $95.82$ & $87.89$ & $91.24$ & $91.68$\\
            Standard deviation & $1.36$ & $7.45$ & $11.56$ & $9.45$ & $9.19$ & $10.15$ & $14.82$ & $15.60$ & $14.21$\\
             &  &  &  &  &  &  &  &  & \\
            UnaryToD\&UnaryDeltaT7 & $98.38$ & $\mathbf{98.95}$ & $98.28$ & $94.32$ & $95.68$ & $95.70$ & $85.96$ & $89.17$ & $90.79$\\
            Standard deviation & $1.31$ & $1.62$ & $4.50$ & $8.93$ & $8.71$ & $6.02$ & $16.23$ & $13.72$ & $14.46$\\
             &  &  &  &  &  &  &  &  & \\
            UnaryToD\&UnaryDeltaT48 & $98.62$ & $98.75$ & $98.66$ & $92.24$ & $\mathbf{96.07}$ & $88.33$ & $88.14$ & $89.48$ & $92.41$\\
            Standard deviation & $1.52$ & $3.07$ & $4.15$ & $11.06$ & $6.35$ & $14.72$ & $16.34$ & $17.32$ & $14.97$\\
             &  &  &  &  &  &  &  &  & \\
            \bottomrule
        \end{tabularx}
    \end{adjustbox}
    \label{acc_houseABC_CRF}
\end{table}

In all the experiments presented in Section~\ref{rep_effect}, the features $hour$ (\textit{ToD}) and $\Delta t$ (\textit{DeltaT}) were not added to the dataset (\textit{NoToD\&NoDeltaT}). In order to test and evaluate the need to better distinguish duration intervals, we considered all the other feature combinations, which result from adding a one-hot(unary-based) encoding of $i$ intervals of the feature $\Delta t$ - $\mathit{OneHotDeltaT_i}$($\mathit{UnaryDeltaT_i}$) - and/or a one-hot(unary-based) encoding of the feature $hour$ - $\mathit{OneHotToD}$ ($\mathit{UnaryToD}$) - to the dataset.

In the following experiment, we test and evaluate the need to better distinguish duration intervals, i.e. the improvements obtained in accuracy by considering more $\Delta t$ values. The results are shown is Table \ref{acc_houseABC_CRF}.

From the results, we see that the best performance for houses A, B and C resulted from the feature combinations \textit{UnaryToD\&UnaryDeltaT7} (5 data points concatenated), \textit{UnaryToD\&UnaryDeltaT48} (5 data points concatenated) and \textit{OneHotToD\&UnaryDeltaT48} (10 data points concatenated), respectively. Furthermore, we observe that only house C significantly benefits from using more $\Delta t$ values and generally, one-hot and unary-based encodings produce similar results for all houses.

Specifically, $98.95 \pm 1.62$ was the best result achieved for house A, where a unary-based encoding with 7 bins was considered. For house B, the best result achieved was $96.07 \pm 6.35$ by applying a unary-based encoding with 48 bins and the best result obtained for house C was $94.10 \pm 15.27$ by using a unary-based encoding with 48 bins.

\subsection{Comparison with State-of-the-art methods}

In this section, we present our best results as well as the corresponding confusion matrices and compare them against the state-of-the-art (Tables~\ref{compsota_houseA}, \ref{compsota_houseB} and \ref{compsota_houseC}). The state-of-the-art methods for this dataset are HSMM and CRF using changepoint and last-fired feature representations, respectively \cite{kasteren11}.

\begin{table}[H]
    \caption{Accuracy and mean per class accuracy rates (\%) and their standard deviation for state-of-the-art methods and our best method for house A - CRF using OB feature representation (UnaryToD\&UnaryDeltaT7 - Data points concatenated: 5)}
    \label{compsota_houseA}
    \scriptsize
    \begin{tabularx}{\columnwidth}{c*{3}{C}}
        \toprule
        Label & HSMM (Changepoint) \cite{kasteren11} & CRF (Last-fired) \cite{kasteren11} & This paper\\
        \midrule
        `Idle' & 50.75 & 86.62 & \textbf{95.98}\\
        `Leave house' & 99.66 & \textbf{99.92} & \textbf{99.92}\\
        `Use toilet' & 82.19 & 61.64 & \textbf{82.74}\\
        `Take shower' & 64.94 & 27.89 & \textbf{82.07}\\
        `Brush teeth' & 34.38 & 0.0 & \textbf{40.62}\\
        `Go to bed' & 96.53 & \textbf{99.76} & 99.64\\
        `Prepare breakfast' & 68.97 & 68.97 & \textbf{86.21}\\
        `Prepare dinner' & 51.57 & 88.85 & \textbf{99.65}\\
        `Get snack' & 54.76 & 14.29 & \textbf{100.0}\\
        `Get drink' & 67.35 & 44.9 & \textbf{89.8}\\
        \midrule
        Mean per class accuracy & 74.96 & 69.35 & \textbf{88.40}\\
        Standard deviation & 12.10 & 12.07 & 12.43\\
        \midrule
        Accuracy & 91.81 & 96.93 & \textbf{98.95}\\
        Standard deviation & 5.88 & 2.11 & 1.62\\
        \bottomrule
    \end{tabularx}
\end{table}

\begin{figure}[H]
    \caption{House A: Confusion matrices of the aforementioned models}
    \centering
    \subfloat[HSMM (Changepoint)]{\label{fig:housea_1}\includegraphics[width=0.32\columnwidth]{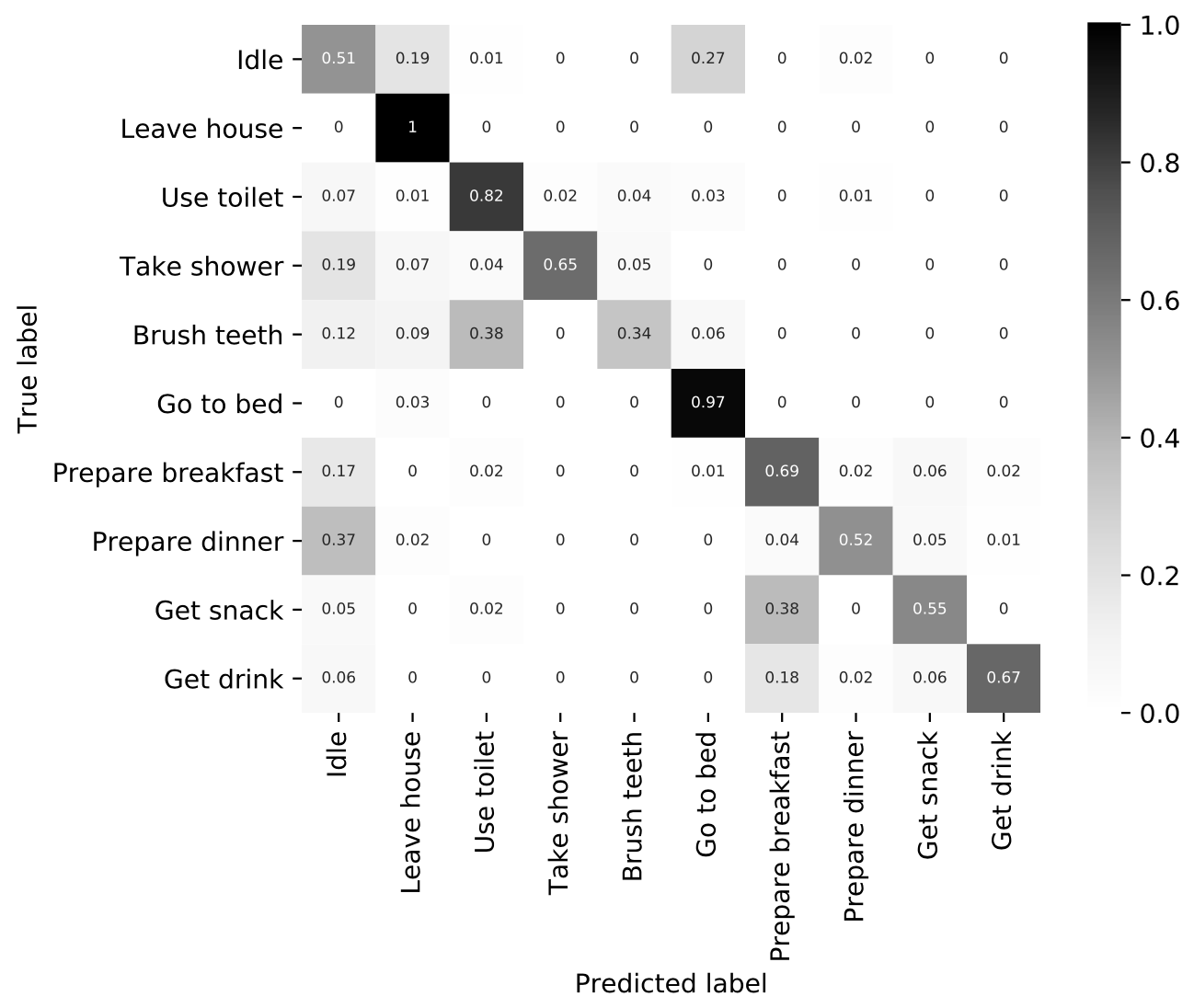}}
    \subfloat[CRF (Last-fired)]{\label{fig:housea_2}\includegraphics[width=0.32\columnwidth]{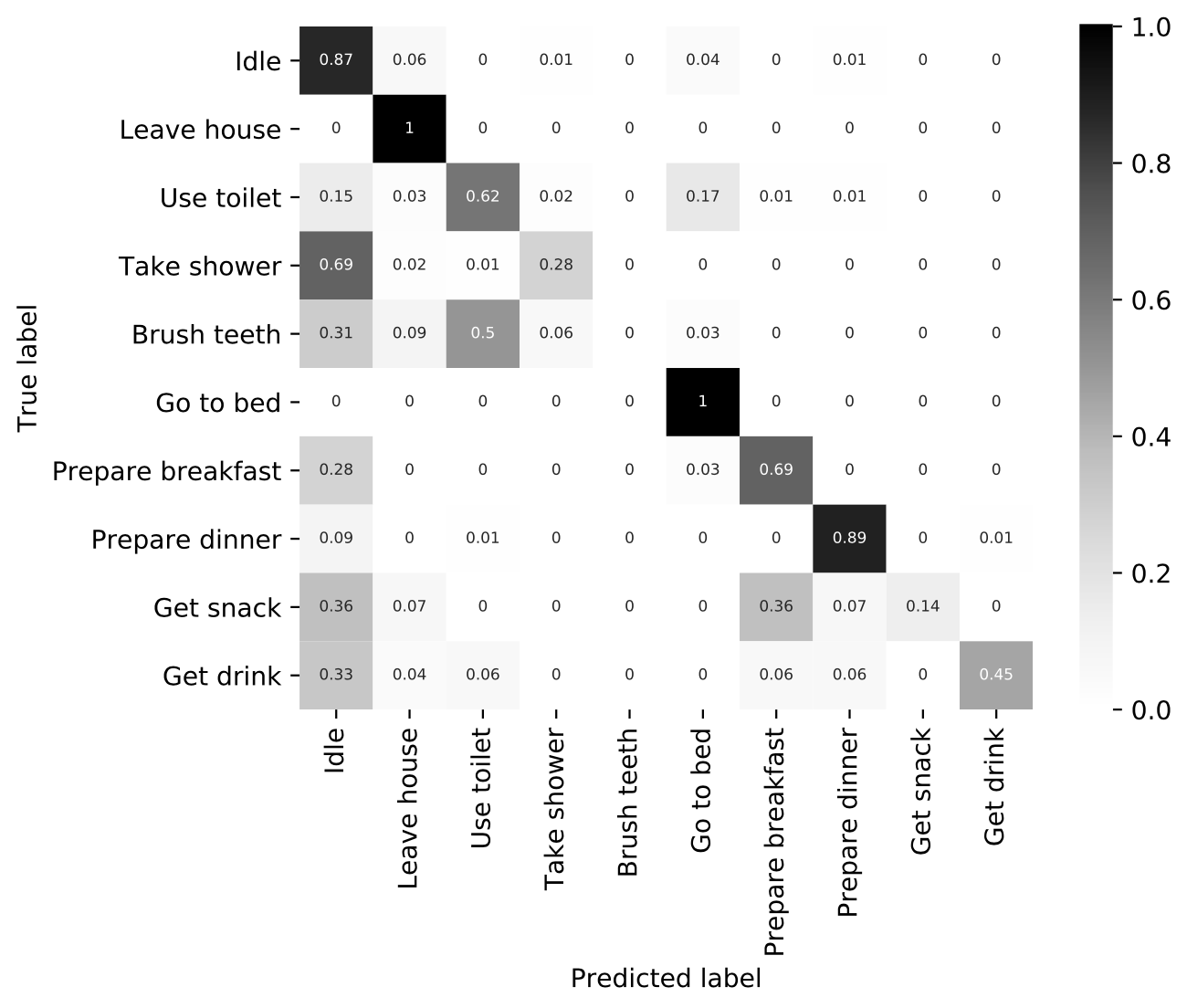}}
    \subfloat[CRF (OB)]{\label{fig:housea_3}\includegraphics[width=0.32\columnwidth]{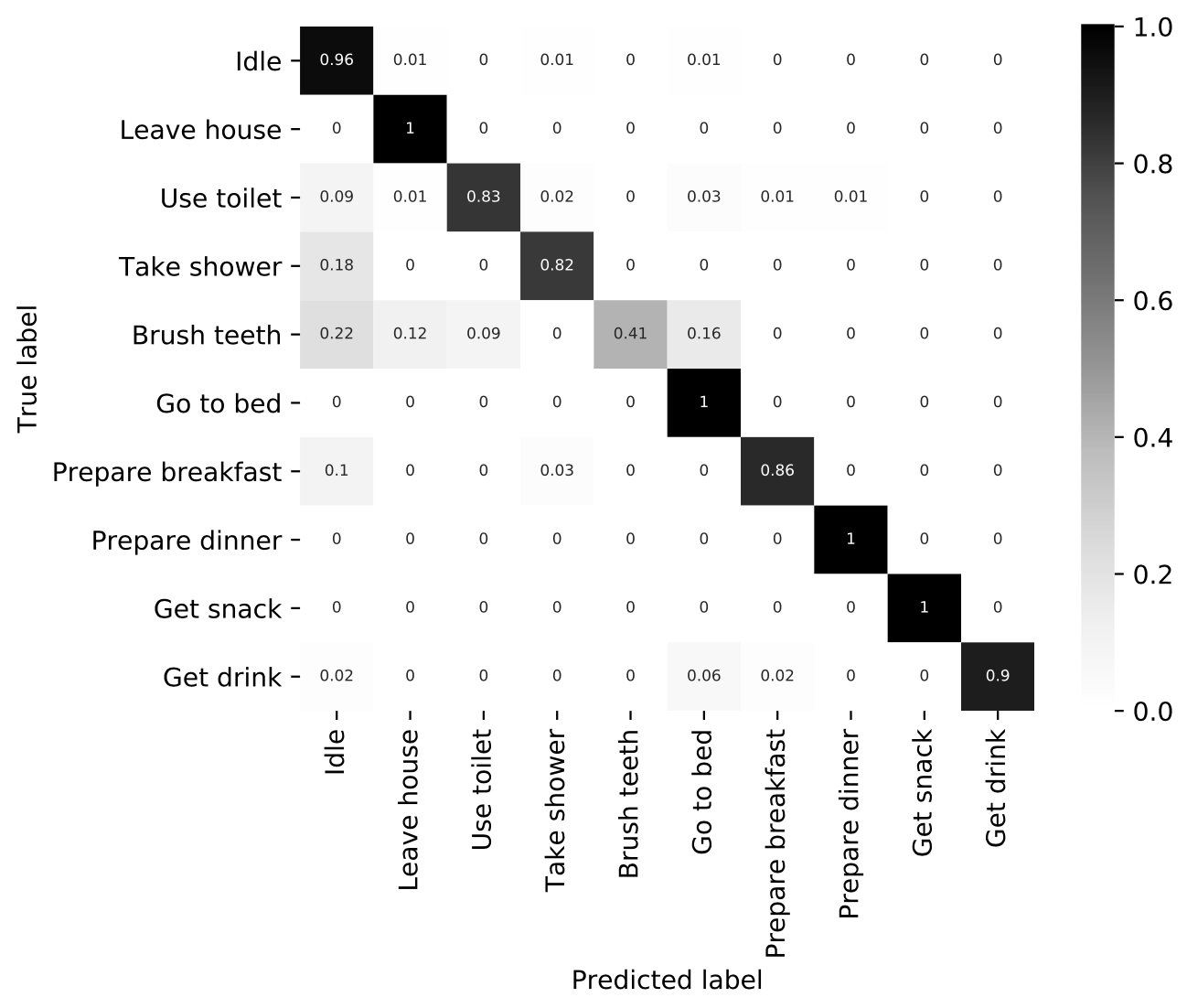}}
    \label{fig:housea_sota_cnfmat}
\end{figure}

For house A (Table~\ref{compsota_houseA}), we observe that the accuracy of every label increased by applying a CRF model with our proposed representation. In particular, the label whose accuracy benefited the most by using the OB representation was `Get snack', which improved by $45\%$. Other labels that had significant improvements were `Take shower' ($17\%$), `Prepare breakfast' ($17\%$) and `Get drink' ($23\%$). On average, considering the label accuracies, we observe an improvement of $13\%$ between the best value obtained from the state-of-the-art methods (HSMM (Changepoint) and CRF (Last-fired)) and the CRF model with our proposed representation.

\begin{table}[H]
    \caption{Accuracy and mean per class accuracy rates (\%) and their standard deviation for state-of-the-art methods and our best method for house B - CRF using OB feature representation (UnaryToD\&UnaryDeltaT48 - Data points concatenated: 5)}
    \label{compsota_houseB}
    \scriptsize
    \begin{tabularx}{\columnwidth}{c*{3}{C}}
        \toprule
        Label & HSMM (Changepoint) \cite{kasteren11} & CRF (Changepoint) \cite{kasteren11} & This paper\\
        \midrule
        `Idle' & 59.86 & 72.62 & \textbf{75.24}\\
        `Leaving the house' & 93.7 & \textbf{99.69} & 99.21\\
        `Use toilet' & \textbf{71.43} & 31.17 & 70.13\\
        `Take shower' & \textbf{92.79} & 87.39 & 72.07\\
        `Brush teeth' & 33.33 & 19.44 & \textbf{63.89}\\
        `Go to bed' & 68.65 & 96.15 & \textbf{97.06}\\
        `Get dressed' & 69.57 & 69.57 & \textbf{86.96}\\
        `Prepare brunch' & 59.52 & 71.43 & \textbf{82.14}\\
        `Prepare dinner' & 38.03 & \textbf{97.18} & 95.77\\
        `Get a drink' & \textbf{42.86} & 14.29 & 28.57\\
        `Wash dishes' & 23.81 & 42.86 & \textbf{71.43}\\
        `Eat dinner' & 42.86 & 0.0 & \textbf{100.0}\\
        `Eat brunch' & 39.04 & 0.0 & \textbf{63.7}\\
        \midrule
        Mean per class accuracy & 65.18 & 58.06 & \textbf{79.08}\\
        Standard deviation & 13.41 & 7.01 & 22.35\\
        \midrule
        Accuracy & 82.27 & 94.99 & \textbf{96.07}\\
        Standard deviation & 13.51 & 5.71 & 6.35\\
        \bottomrule
    \end{tabularx}
\end{table}

\begin{figure}[H]
    \caption{House B: Confusion matrices of the aforementioned models}
    \centering
    \subfloat[HSMM (Changepoint)]{\label{fig:houseb_1}\includegraphics[width=0.32\columnwidth]{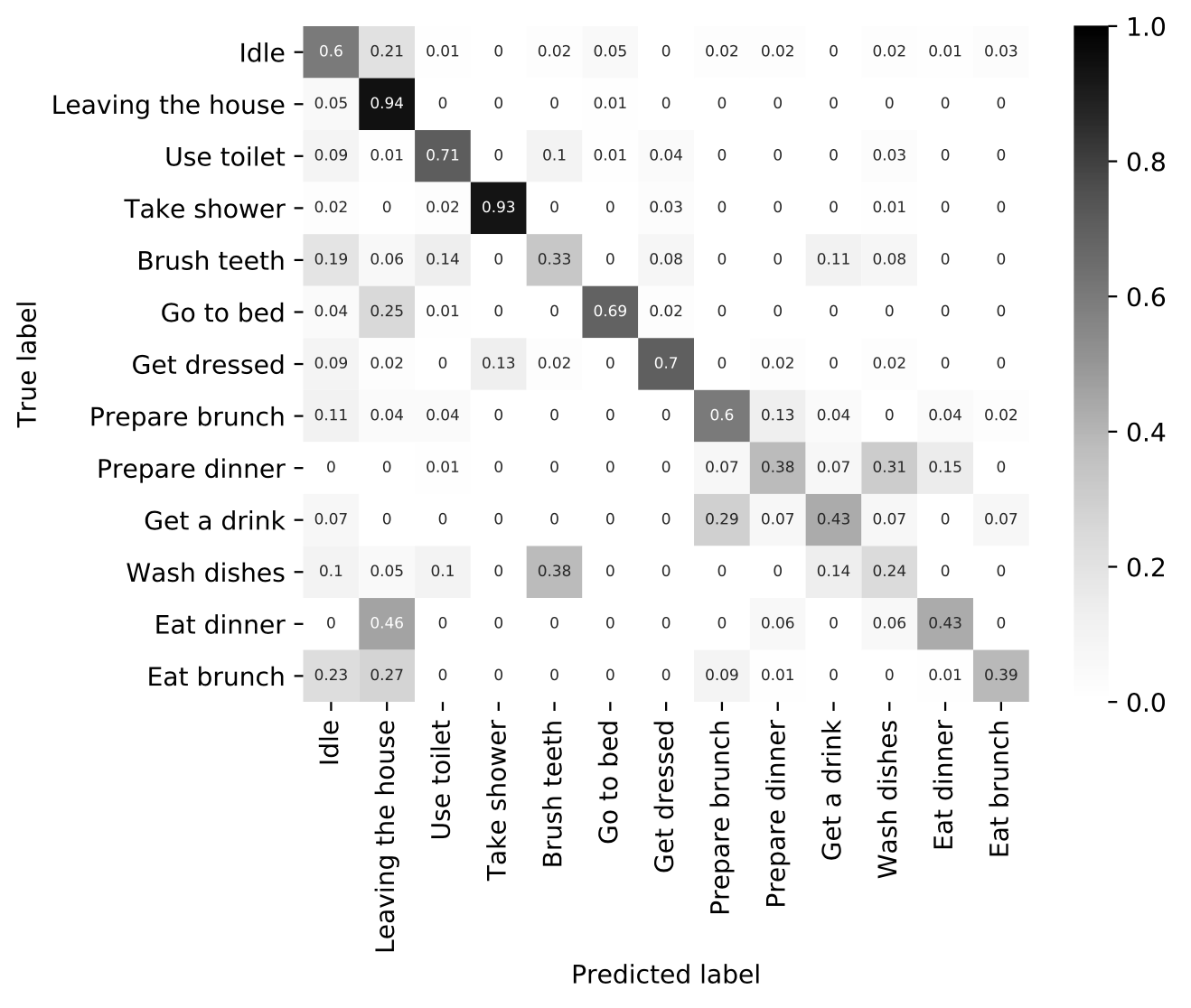}}
    \subfloat[CRF (Changepoint)]{\label{fig:houseb_2}\includegraphics[width=0.32\columnwidth]{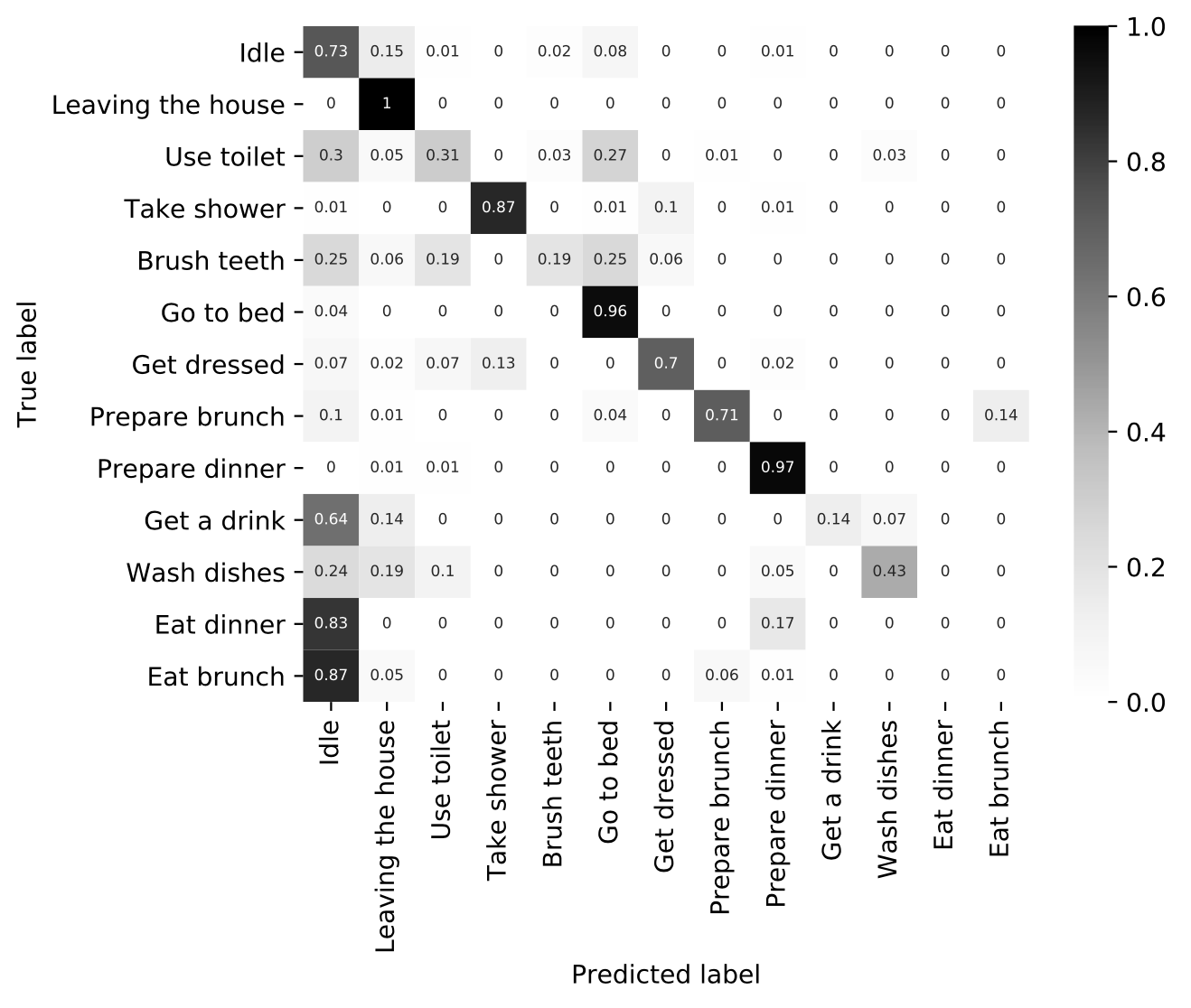}}
    \subfloat[CRF (OB)]{\label{fig:houseb_3}\includegraphics[width=0.32\columnwidth]{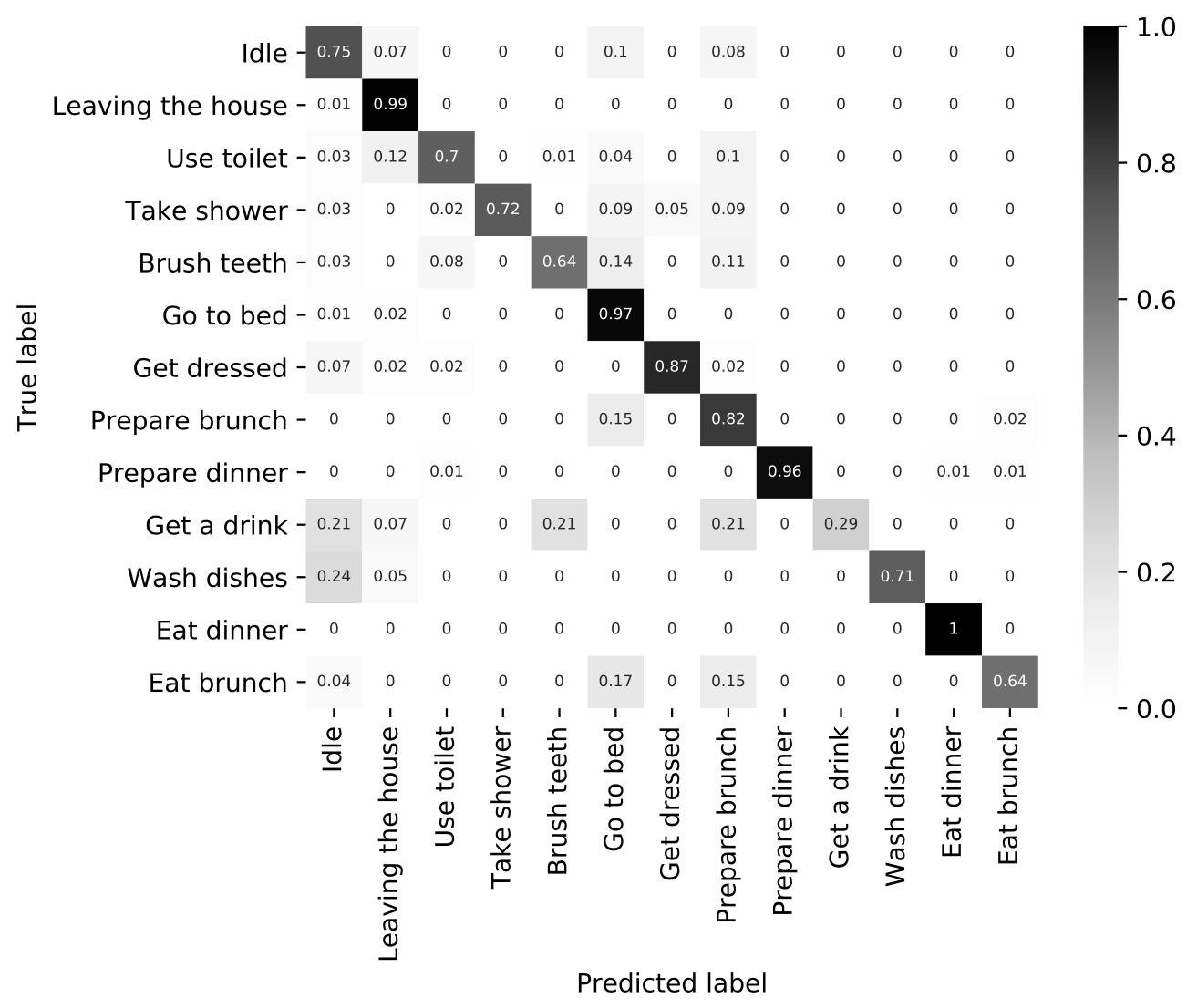}}
    \label{fig:houseb_sota_cnfmat}
\end{figure}

In regard to house B (Table~\ref{compsota_houseB}), we observe that the accuracy of most of the labels improves, but the labels `Take shower' and `Get a drink' decrease by $21\%$ and $14\%$, respectively. In particular, the accuracy of label `Take shower' decreases due to being misclassified as `Going to bed' and `Prepare brunch'. As for label `Get a drink', it is classified $63\%$ of the times as `Idle', `Brush teeth' and `Prepare brunch'. Nevertheless, on average, we obtain an improvement of $10.3\%$ between the best value obtained from the state-of-the-art methods and the CRF model with the OB representation. 

\begin{table}[H]
    \caption{Accuracy and mean per class accuracy rates (\%) and their standard deviation for state-of-the-art methods and our best method for house C - CRF using OB feature representation (OneHotToD\&UnaryDeltaT48 - Data points concatenated: 10)}
    \label{compsota_houseC}
    \scriptsize
    \begin{tabularx}{\columnwidth}{c*{3}{C}}
        \toprule
        Label & HSMM (Last-fired) \cite{kasteren11} & CRF (Last-fired) \cite{kasteren11} & This paper\\
        \midrule
        `Idle' & 68.57 & 82.6 & \textbf{85.81}\\
        `Leave house' & 86.19 & 95.96 & \textbf{98.14}\\
        `Eating' & 22.19 & 6.73 & \textbf{72.07}\\
        `Use toilet downstairs' & \textbf{63.29} & 21.52 & 27.85\\
        `Take shower' & 60.0 & 36.32 & \textbf{81.58}\\
        `Brush teeth' & 26.73 & 4.95 & \textbf{78.22}\\
        `Use toilet upstairs' & 45.0 & 13.75 & \textbf{52.5}\\
        `Shave' & 43.48 & 31.88 & \textbf{97.1}\\
        `Go to bed' & 98.03 & \textbf{99.37} & 96.76\\
        `Get dressed' & 69.64 & 56.25 & \textbf{81.25}\\
        `Take medication' & 26.67 & 0.0 & \textbf{40.0}\\
        `Prepare breakfast' & 33.8 & 49.3 & \textbf{76.06}\\
        `Prepare lunch' & 48.33 & 41.67 & \textbf{83.33}\\
        `Prepare dinner' & 69.31 & 55.86 & \textbf{90.69}\\
        `Get snack' & 20.83 & 4.17 & \textbf{66.67}\\
        `Get drink' & 0.0 & 6.45 & \textbf{51.61}\\
        \midrule
        Mean per class accuracy & 55.98 & 46.79 & \textbf{76.54}\\
        Standard deviation & 15.4 & 15.63 & 18.99\\
        \midrule
        Accuracy & 84.48 & 90.69 & \textbf{94.10}\\
        Standard deviation & 13.17 & 9.05 & 15.27\\
        \bottomrule
    \end{tabularx}
\end{table}

\begin{figure}[H]
    \caption{House C: Confusion matrices of the aforementioned models}
    \centering
    \subfloat[HSMM (Last-fired)]{\label{fig:housec_1}\includegraphics[width=0.32\columnwidth]{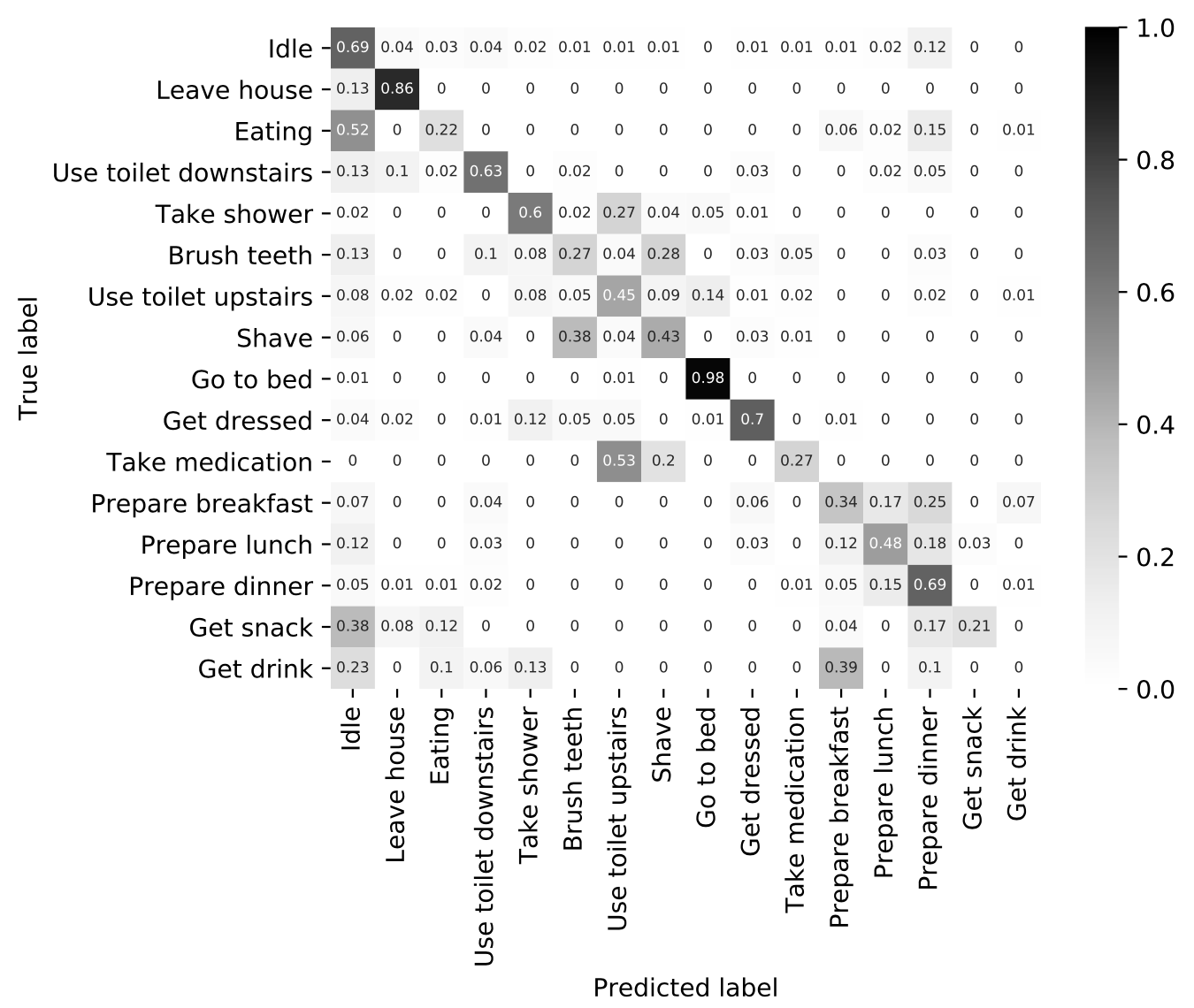}}
    \subfloat[CRF (Last-fired)]{\label{fig:housec_2}\includegraphics[width=0.32\columnwidth]{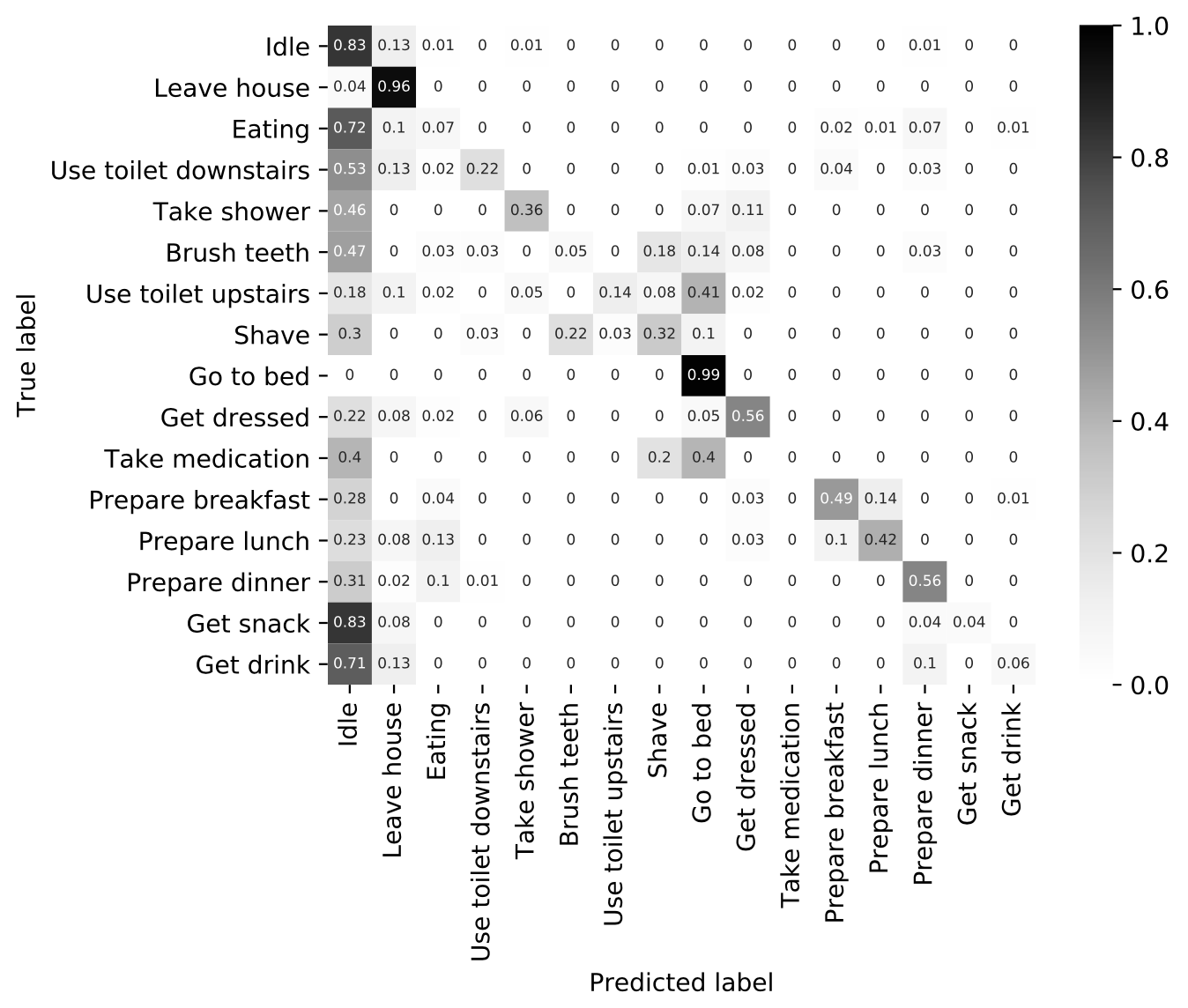}}
    \subfloat[CRF (OB)]{\label{fig:housec_3}\includegraphics[width=0.32\columnwidth]{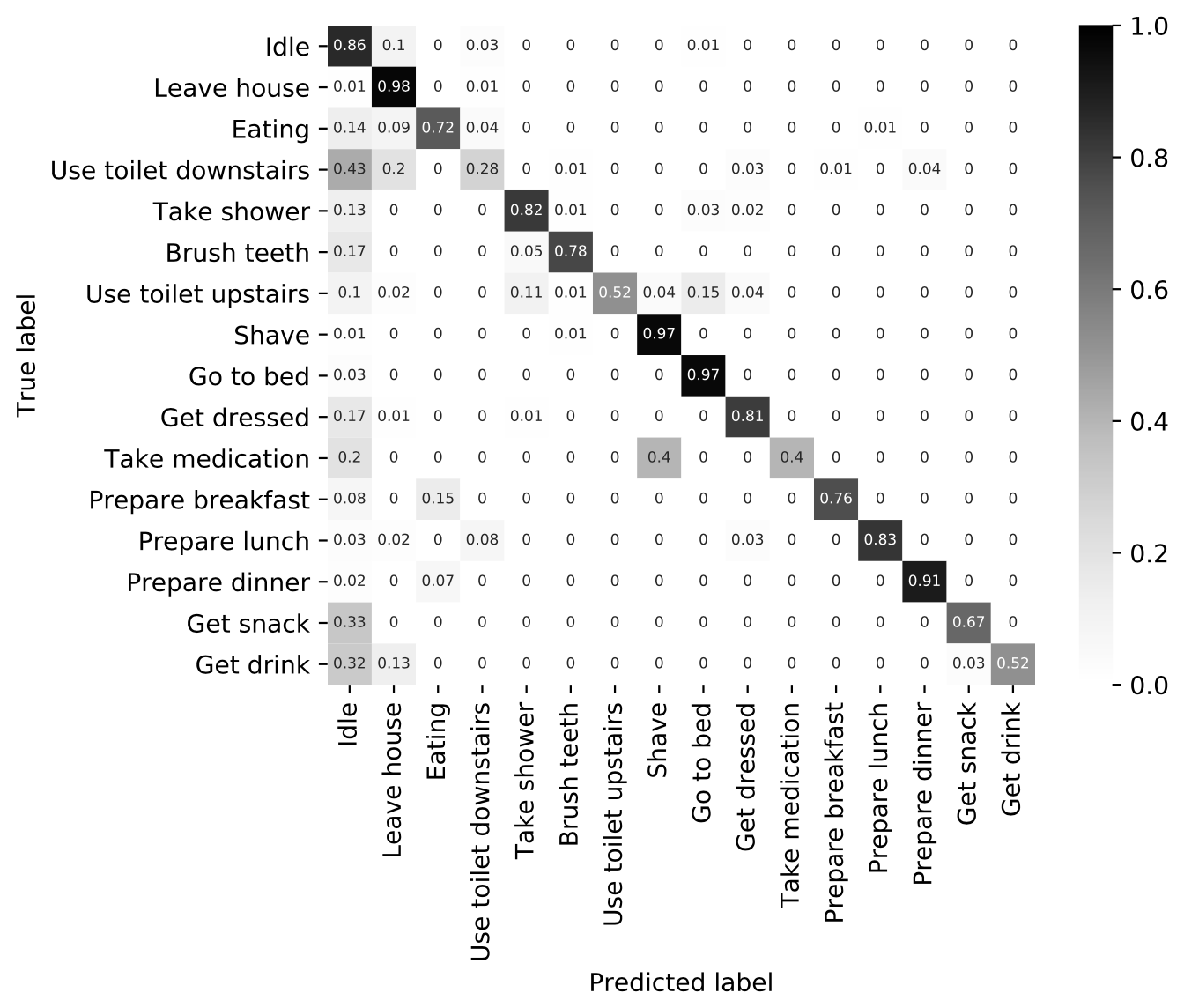}}
    \label{fig:housec_sota_cnfmat}
\end{figure}

\pagebreak
We observe that the largest improvement regarding label accuracy was given by house C (Table~\ref{compsota_houseC}): on average, there was an improvement of $22\%$ between the best value obtained from the state-of-the-art methods (HSMM (Changepoint) and CRF (Last-fired)) and the CRF model with our proposed representation. One exception we observe is the label `Use toilet downstairs'. The highest accuracy for this label is obtained with the HSMM method and a last-fired representation. This occurs because, most of the times, the other two feature representations misclassify this highly infrequent label as `Idle' (see confusion matrices in Figures \ref{fig:housec_2} and \ref{fig:housec_3}). From Figure \ref{fig:housec}, we know that this is a highly infrequent label in this dataset.

From the experiments above, we conclude that the OB representation outperformed the state-of-the-art feature representations and, in general, there is not only a significant improvement in the accuracies for each class but also in the overall accuracy.

Even though CRFs outperform HSMM from an overall accuracy standpoint, when considering the per-class accuracy, HSMMs are sometimes able to better classify infrequent classes in comparison to CRFs. This results from the learning process each method is undertaking. Specifically, HSMMs build a model $p(\mathbf{x_{t}}|\mathbf{y_{t}})$ for each class, whereas CRFs use the same model for all classes by computing $p(\mathbf{c}|\mathbf{X})$, which causes competition among classes. Consequently, if a dataset is imbalanced, a higher likelihood may be obtained if the data points are classified as the dominant class(es) than if the low frequent classes are considered and some of the dominant ones are misclassified \cite{kasteren08}.

%% file: conclusion.tex
\section{Conclusion} \label{sec_5}

In this paper, we have presented a thorough study of different ML techniques for a standard HAR dataset. Our experiments show that a significant improvement was made in comparison to state-of-the-art methods in the HAR field.

A new representation for data that is to be given as input to a model was presented. The results have shown that, by applying such a representation, models are better able to learn data patterns and, consequently, successfully perform a classification task in the HAR domain for both dominant and minor classes.

By using an OB representation, we improved the mean per-class accuracy and the accuracy for house A by $13.44\%$ and $2.02\%$, respectively, in comparison with the state-of-the-art results. Moreover, for house B, the aforementioned evaluation metrics increased $13.9\%$ and $1.08\%$, respectively. As for house C, results improved $20.56\%$ and $3.41\%$ for the respective evaluation metrics considered.

Given the results obtained with an observation-based representation, its usage may also be suitable and advantageous in other domains. Moreover, using adversarial zero-shot learning \cite{lampert09, tong18} to recognise abnormal human activity  is an interesting direction for future work.

%% file: appendix.tex
\section{Appendix} \label{appendix}

Below is a list of the results that we have compiled based on the literature with regard to the three datasets (Houses A, B and C).

\begin{table}[H]
    \scriptsize
    \caption{Results obtained for Houses A, B and C}
    \begin{adjustbox}{width=\columnwidth,center}
        \begin{tabularx}{\textwidth}{c *{8}{C}} 
            \toprule
            \multicolumn{1}{c}{Model} & \multicolumn{1}{c}{Feature} & \multicolumn{3}{c}{Mean per class accuracy} & \multicolumn{3}{c}{Accuracy}\\
            \cmidrule(l){3-5} \cmidrule(l){6-8}
             & \multicolumn{1}{c}{Representation} & A & B & C & A & B & C\\ 
            \midrule
            NB \cite{kasteren11} & Raw & $42.6$ & $32.5$ & $16.8$ & $77.1$ & $80.4$ & $46.5$\\ 
            
            NB \cite{kasteren11} & Changepoint & $43.2$ & $38.9$ & $30.8$ & $55.9$ & $67.8$ & $57.6$\\
            
            NB \cite{kasteren11} & \mbox{Last-fired} & $64.8$ & $44.6$ & $46.4$ & $95.3$ & $86.2$ & $87.0$\\ 
            \midrule
            HMM \cite{kasteren11} & Raw & $45.5$ & $44.7$ & $17.2$ & $59.1$ & $63.2$ & $26.5$\\ 
            
            HMM \cite{kasteren11} & Changepoint & $74.3$ & $63.1$ & $50.0$ & $92.3$ & $81.0$ & $77.2$\\
            
            HMM \cite{kasteren11} & \mbox{Last-fired} & $69.5$ & $46.6$ & $53.7$ & $89.5$ & $48.4$ & $83.9$\\ 
            \midrule
            HSMM \cite{kasteren11} & Raw & $48.5$ & $44.6$ & $20.4$ & $59.5$ & $63.8$ & $31.2$\\
            
            HSMM \cite{kasteren11} & Changepoint & $75.0$ & $65.2$ & $52.3$ & $91.8$ & $82.3$ & $77.5$\\
            
            HSMM \cite{kasteren11} & \mbox{Last-fired} & $73.8$ & $53.3$ & $56.0$ & $91.0$ & $67.1$ & $84.5$\\ 
            \midrule
            CRF \cite{kasteren11} & Raw & $56.1$ & $40.6$ & $21.8$ & $89.8$ & $78.0$ & $46.3$\\ 
            
            CRF \cite{kasteren11} & Changepoint & $68.0$ & $51.5$ & $39.6$ & $91.4$ & $92.9$ & $82.2$\\
            
            CRF \cite{kasteren11} & \mbox{Last-fired} & $65.8$ & $47.8$ & $40.4$ & $96.4$ & $89.2$ & $89.7$\\ 
            \midrule
            Vanilla \cite{arifoglu17} & Raw & $64.8$ & $46.9$ & $43.1$ & $86.8$ & $65.2$ & $50.2$\\ 
            
            Vanilla \cite{arifoglu17} & Changepoint & $63.8$ & $62.4$ & $54.9$ & $61.4$ & $76.9$ & $72.2$\\
            
            Vanilla \cite{arifoglu17} & \mbox{Last-fired} & $74.3$ & $64.4$ & $59.6$ & $95.5$ & $87.9$ & $86.7$\\ 
            \midrule
            LSTM \cite{arifoglu17} & Raw & $63.9$ & $44.0$ & $34.8$ & $86.7$ & $63.5$ & $45.3$\\ 
            
            LSTM \cite{arifoglu17} & Changepoint & $63.6$ & $59.0$ & $53.3$ & $61.4$ & $76.8$ & $72.0$\\
            
            LSTM \cite{arifoglu17} & \mbox{Last-fired} & $73.9$ & $60.1$ & $57.3$ & $96.7$ & $87.2$ & $87.4$\\ 
            \midrule
            GRU \cite{arifoglu17} & Raw & $69.1$ & $36.3$ & $33.2$ & $86.6$ & $64.5$ & $46.7$\\ 
            
            GRU \cite{arifoglu17} & Changepoint & $65.0$ & $53.5$ & $47.0$ & $61.4$ & $76.4$ & $71.6$\\
            
            GRU \cite{arifoglu17} & \mbox{Last-fired} & $80.6$ & $56.9$ & $52.7$ & $96.1$ & $87.0$ & $86.6$\\ 
            \midrule
            SVM \cite{arifoglu17} & Raw & $69.1$ & $58.5$ & $35.2$ & $85.4$ & $81.6$ & $37.4$\\ 
            
            SVM \cite{arifoglu17} & Changepoint & $63.4$ & $53.6$ & $51.4$ & $55.9$ & $67.9$ & $57.8$\\
            
            SVM \cite{arifoglu17} & \mbox{Last-fired} & $77.2$ & $54.6$ & $55.5$ & $96.1$ & $86.2$ & $87.5$\\
            \midrule
            LSTM \cite{singh17} & Raw & - & - & - & $89.8$ & $85.7$ & $64.22$\\
            
            LSTM \cite{singh17} & \mbox{Last-fired} & - & - & - & $95.3$ & $88.5$ & $85.9$\\ 
            \bottomrule
        \end{tabularx}
    \end{adjustbox}
    \label{sotares_table}
\end{table}